\documentclass[aip, floatfix, jcp, reprint, longbibliography, nofootinbib]{revtex4-1}

\usepackage[utf8]{inputenc}
\usepackage{amsmath}
\usepackage{amssymb}
\usepackage[nopar]{lipsum}
\usepackage{mathtools}
\usepackage{nameref,hyperref}
\usepackage{microtype}
\usepackage{color}
\usepackage{graphicx}
\usepackage{soul}
\usepackage{appendix}
\usepackage{booktabs}
\usepackage{float}
\usepackage{subfig}

\usepackage{hyperref}

\renewcommand{\thefootnote}{\roman{footnote}}

\begin{document}

\makeatletter
\let\@fnsymbol\@fnsymbol@latex
\@booleanfalse\altaffilletter@sw
\def\@fnsymbol#1{\ensuremath{\ifcase#1\or \dagger\or *\or \ddagger\or
   \mathsection\or \mathparagraph\or \|\or **\or \dagger\dagger
   \or \ddagger\ddagger \else\@ctrerr\fi}}
\makeatother
\renewcommand{\thefootnote}{\fnsymbol{footnote}}

\title{Step Change Improvement in ADMET Prediction with \\ PotentialNet Deep Featurization}

\author{Evan N. Feinberg}
\email{evan.n.feinberg@gmail.com}
\altaffiliation{\\Currently at: Genesis Therapeutics, Inc.}
\affiliation{Program in Biophysics, Stanford University, Palo Alto, CA, USA}
\affiliation{Computational and Structural Chemistry, Merck \& Co., Inc., South San Francisco, CA, USA}
\author{Robert Sheridan}
\affiliation{Computational and Structural Chemistry, Merck \& Co., Inc., Kenilworth, NJ, USA}
\author{Elizabeth Joshi}
\affiliation{Pharmacokinetics, Pharmacodynamics, and Drug Metabolism, Merck \& Co., Inc., Kenilworth, NJ, USA}
\author{Vijay S. Pande}
\affiliation{Department of Bioengineering, Stanford University, Palo Alto, CA, USA}
\author{Alan C. Cheng}
\email{alan.cheng@merck.com}
\affiliation{Computational and Structural Chemistry, Merck \& Co., Inc., South San Francisco, CA, USA}

% \begin{abstract}
% \end{abstract}

\maketitle
\renewcommand{\thefootnote}{\roman{footnote}}
\section*{Abstract}

The Absorption, Distribution, Metabolism, Elimination, and Toxicity (ADMET) properties of drug candidates are estimated to account for up to 50\% of all clinical trial failures \cite{kennedy1997managing, kola2004can}. Predicting ADMET properties has therefore been of great interest to the cheminformatics and medicinal chemistry communities in recent decades. Traditional cheminformatics approaches, whether the learner is a random forest or a deep neural network, leverage fixed fingerprint feature representations of molecules. In contrast, in this paper, we learn the features most relevant to each chemical task at hand by representing each molecule explicitly as a graph, where each node is an atom and each edge is a bond. By applying graph convolutions to this explicit molecular representation, we achieve, to our knowledge, unprecedented accuracy in prediction of ADMET properties. By challenging our methodology with rigorous cross-validation procedures and prospective analyses, we show that deep featurization better enables molecular predictors to not only interpolate but also extrapolate to new regions of chemical space.

\section{Introduction}
Only about 12\% of drug candidates entering human clinical trials ultimately reach FDA approval\cite{kola2004can}. This low success rate stems to a significant degree from issues related to the absorption, distribution, metabolism, elimination, and toxicity (ADMET) properties of a molecule. In turn, ADMET properties are estimated to account for up to 50\% of all clinical trial failures \cite{kennedy1997managing, kola2004can}. 

Over the past few years, Merck\footnote{Merck \& Co., Inc., Kenilworth, NJ, USA, in the United States; MSD internationally} has been heavily invested in leveraging institutional knowledge in an effort to drive hypothesis-driven, model-guided experimentation early in discovery.  To that end, \textit{in silico} models have been established for many of our early screening assay endpoints deemed critical in the design of suitable potential candidates in order to selectively invest available resources in chemical matter having the best possible chance of delivering an efficacious and safe drug candidate in a timely fashion~\cite{sherer2012qsar, sanders2017informing}.

Supervised machine learning (ML) is an umbrella term for a family of functional forms and optimization schemes for mapping input features representing input samples to ground truth output labels. The traditional paradigm of ML involves representing training samples as flat vectors of features \cite{hastie2009overview}. This \textit{featurization} step frequently entails domain-specific knowledge. For instance, recent work in protein-ligand binding affinity prediction represents, or featurizes, a protein-ligand co-crystal complex with a flat vector containing properties including number of hydrogen bonds, number of salt bridges, number of ligand rotatable bonds, and floating point measures of such properties as hydrophobic interactions \cite{durrant2011nnscore, ballester2010machine, li2015improving}.

In the domain of ligand-based QSAR, cheminformaticians have devised a variety of schemes to represent individual molecules as flat vectors of mechanized features. Circular fingerprints \cite{rogers2010extended} of bond diameter 2, 4, and 6 (ECFP2, ECFP4, ECFP6, respectively) hash local neighborhoods of each atom to bits in a fixed vector. In contrast, atom pair features \cite{carhart1985atom} denote pairs of atoms in a given molecule, the atom types, and the minimum bond path length that separates the two atoms. In a supervised ML setting, regardless of specific featurization, all such fixed length vector featurizations will be paired with a learning algorithm of choice (e.g., Random Forests, Support Vector Machines\cite{hastie2009overview}, multilayer perceptron deep neural networks, i.e. MLP DNN's \cite{goodfellow2016deep}) that will then return a mapping of features to some output assay label of interest. 

While circular fingerprints, atom pair features, MACCS keys \cite{durant2002reoptimization}, and related generic schemes have the potential to supersede the signal present in hand-crafted features, they are still inherently limited by the insuperable inefficiency of projecting a complex multidimensional object onto a single dimension. Whereas graph convolutions win with molecules by exploiting the concept of bond adjacency, two-dimensional convolutions win with images by exploiting pixel adjacency, and recurrent neural networks win by exploiting temporal adjacency, there is no meaning to proximity of bits along either an ECFP or pair descriptor set of molecular features. For instance, considering the pair descriptor framework, an $sp^3$ carbon that is five bonds away from an $sp^2$ nitrogen might denote the first bit in the feature vector, an $sp^2$ oxygen that is two bonds away from an $sp^3$ nitrogen might denote the very next bit in the same feature vector, and an $sp^3$ carbon that is four bonds away from an $sp^2$ nitrogen might denote the hundredth bit in the feature vector. Put in descriptor language, a feature like ``CX3sp3-04-NX1sp2'' is conceptually similar to ``CX20sp3-04-NX2sp2'', but the descriptors are treated as fully unrelated in descriptor-based QSAR while graph convolutional approaches could separate the ``element'' component from the ``hybridization'' component and both from the ``bond distance'' component. The conceptual proximity between qualitatively similar descriptors is weakened by the arbitrary arrangement of bits in the featurization process and therefore must be ``re-learned'' by the supervised machine learning algorithm of choice.

\begin{figure*}[!htb]%[H]
\caption{Comparison of algorithms for multilayer perceptron versus graph convolution.}
\label{mlp_vs_gcnn}
\includegraphics[width=6in]{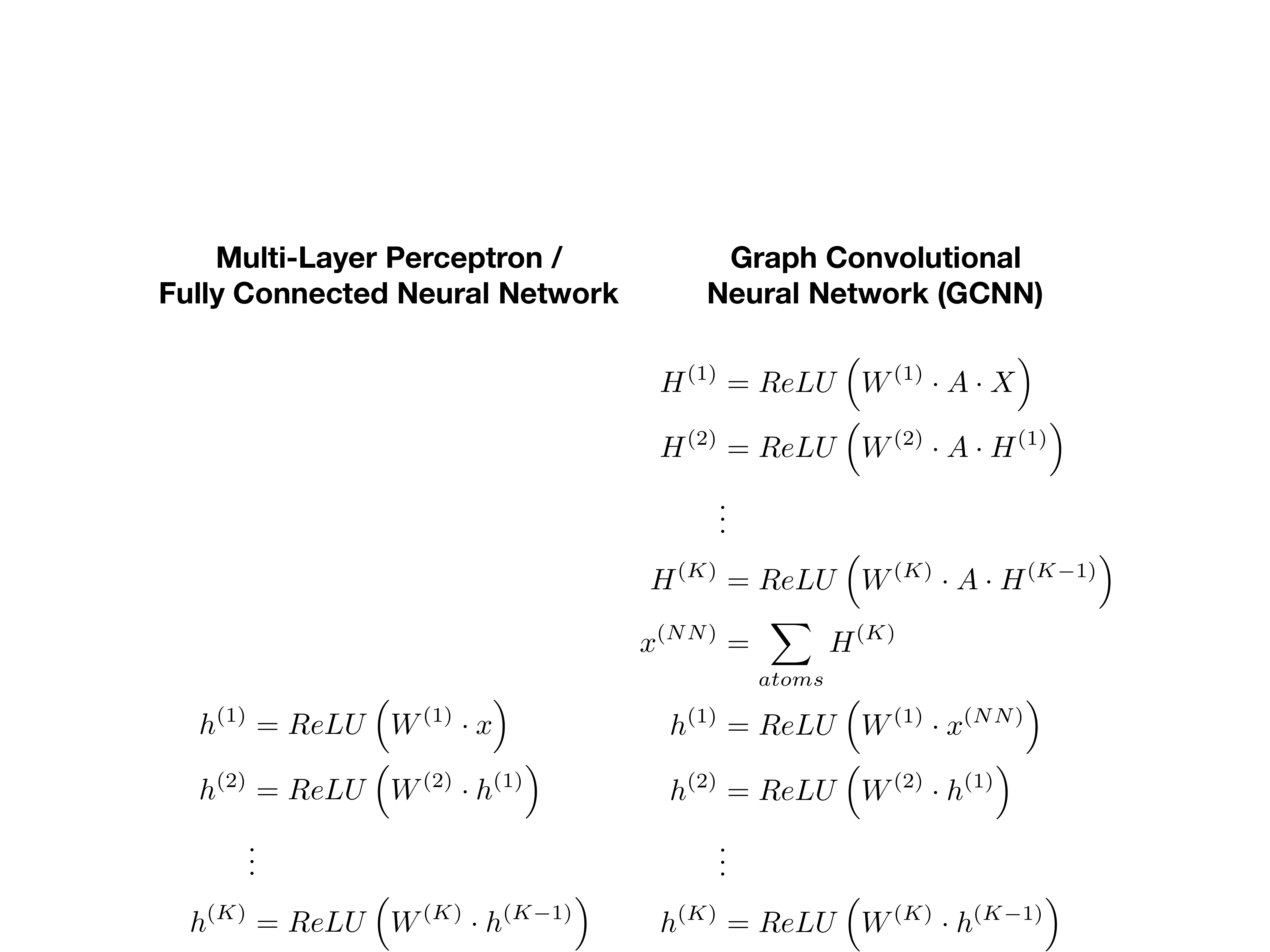}
\end{figure*}

To both compare and contrast chemical ML on fixed vector descriptors with chemical deep learning on graph features, we write out a multilayer perceptron (MLP) and a graph convolutional neural network (GCNN) side-by-side (Figure \ref{mlp_vs_gcnn}). In Figure \ref{mlp_vs_gcnn}, each molecule is represented either by a flat vector $x$ for the MLP or by both an $N_{atoms} \times N_{atoms}$ adjacency matrix $A$ and an $N_{atoms} \times f_{in}$ per-atom feature matrix $X$ for the GCNN. The GCNN begins with $K$ graph convolutional layers. It then proceeds to a graph gather operation that sums over the per-atom features in the last graph convolutional hidden layer: $x^{(NN)} = \sum_{atoms} H^{(K)}$, where the differentiable $x^{(NN)}$, by analogy to the fixed input $x$ of the MLP, is a flat vector graph convolutional fingerprint for the entire molecule, and $H^{(K)}_i$ is the feature map at the $K'th$ graph convolutional layer for atom $i$. The final layers of the GCNN are identical in form to the hidden layers of the MLP. The difference, therefore, between the MLP and the GCNN lies in the fact that $x$ for MLP is a fixed vector of molecular fingerprints, whereas $x^{(NN)}$ of the GCNN is an end-to-end differentiable fingerprint vector: the features are \textit{learned} in the graph convolution layers. 

Another noteworthy parallel arises between the MLP hidden layers and the GCNN graph convolutional layers. Whereas the first MLP layer maps $h^{(1)} = ReLU(W^{(1)} \cdot x)$, the GCNN inserts the adjacency matrix $A$ between $W$ and atom feature matrix $X$: $H^{(1)} = ReLU(W^{(1)} \cdot A \cdot X)$. Note that, while the feature maps $X, H^{(1)}, ..., H^{(K)}$ change at each layer of a GCNN, the adjacency matrix $A$ is a constant to be re-used at each layer. Therefore, in a recursive manner, a given atom is passed information about other atoms succesively further in bond path length at each graph convolutional layer. 

Since the advent of the basic graph convolutional neural network, a spate of new approaches\cite{kearnes2016molecular, kipf2016semi, li2015gated, gilmer2017neural, feinberg2018potentialnet} have improved upon the elementary graph convolutional layers expressed in Figure \ref{mlp_vs_gcnn}. Here, we train neural networks based on the PotentialNet \cite{feinberg2018potentialnet} family of graph convolutions.

\begin{equation}\label{S-GGNN}
\begin{aligned}
 h^{(1)}_i &= GRU \left (x_i, \sum^{N_{\text{et}}}_e \sum_{j \in N^{(e)}(v_i)} NN^{(e)}  (x_j) \right ) \\ 
 &\vdots& \\
 h^{(K)}_i &= GRU \left (h^{(b_{K-1})}_i, \sum^{N_{\text{et}}}_e \sum_{j \in N^{(e)}(v_i)} NN^{(e)}  (h^{(b_{K-1})}_j) \right ) \\
 h^{(NN)} &= \sigma \left (i(h^{(K)}, x) \right ) \odot  \left (j (h^{(K)}) \right ) \\ 
 h^{(FC_0)} &= \sum^{N_{Lig}}_{j=1} h^{(NN)}_j \\
h^{(FC_1)} &= ReLU \left ( W^{(FC_1)} h^{(FC_0)} \right ) \\
&\vdots& \\
h^{(FC_K)} &= W^{(FC_K)} h^{(FC_{K-1})}
\end{aligned}
\end{equation}

\noindent{} where $h_i^{(k)}$ represents the feature map for $atom_i$ at graph convolutional layer $k$; $i$, $j$, and $NN$ are neural networks, $N_{Lig}$ is the number of ligand atoms, $\{W\}$ are weight matrices for different layers. The GRU is a Gated Recurrent Unit which affords a more efficient passing of information to an atom from its neighbors. In this way, whereas one might view the standard GCNN displayed in Figure \ref{mlp_vs_gcnn} as more analogous to learnable and more efficient ECFP featurization, one might view the PotentialNet layers in Equation \ref{S-GGNN} as being more analogous to a learnable and more efficient pair descriptor\cite{carhart1985atom} featurization.

In this paper, we conduct a direct comparison of the current state-of-the-art algorithm similar to those used by many major pharmaceutical companies -- random forests based on atom pair descriptors -- with PotentialNet (Equation \ref{S-GGNN}). We trained ML models on thirty-one chemical datasets describing results of various ADMET assays -- ranging from physicochemical properties to animal-based PK properties -- and compared results of random forests with those of PotentialNet on held out test sets chosen by two different cross-validation strategies (further details on PotentialNet training are included in Methods). In addition, we ascertain the capacity of our trained models to generalize to assays conducted outside of our institution by downloading datasets from the scholarly literature and conducting further performance comparisons. Finally, we conduct a prospective comparison of prediction accuracy of Random Forests and PotentialNet on assay data of new chemical entities recorded well after all model parameters were frozen in place.

%In the following sections, we conduct an extensive comparison of performance of 

%While MLP type neural networks can outperform their predecessor methods, like Random Forests and SVMs, when trained on the same sets of features, deep neural networks began to make their mark with the advent of the convolutional neural network. 

%Deep neural networks (DNN) denote a set of functional forms which frequently surpass previous generations of ML methods by learning the features pertinent to the prediction task at hand in intermediate neural network layers. Traditional cheminformatics approaches, whether the learner is a random forest or a DNN, leverage fixed fingerprint feature representations of molecules (cite). In contrast, in this paper, we ``learn'' the features most relevant to each chemical task at hand by representing each molecule explicitly as a graph, where each node is an atom and each edge is a bond. By applying graph convolutions to this explicit molecular representation, we achieve unprecedented accuracy in prediction of ADMET properties. By challenging our methodology with difficult cross-validation procedures, we show that deep featurization better enables molecular predictors to not only interpolate, but to extrapolate to new regions of chemical space.

\section{Results} 

The primary purpose of supervised machine learning is to train computer models to make accurate predictions about samples that have not been seen before in the learning process. In the discipline of computer vision, the ability to interpolate between training samples is often sufficient for real-world applications. Such is not the case in the field of medicinal chemistry. When a chemist is tasked with generating new molecular entities to selectively modulate a given biological target, they must invent chemical matter that is fundamentally different than previously known materials. This need stems from both scientific and practical concerns; every biological target is different and likely requires heretofore nonexistent chemical matter, and the patent system demands that, to garner protection, new chemical entities must be not only useful but also novel and sufficiently different from currently existing molecules.

Cross-validation is a subtle yet critical component of any ML initiative. In the absence of the ability to gather prospective data, it is standard practice in ML to divide one's retrospectively available training data into three disjoint subsets: train, valid, and test (though it is only strictly necessary that the test set be disjoint from the others). It is well known that cross-validation strategies typically used in the vision or natural language domains, like random splitting, significantly over-estimate the generalization and extrapolation ability of machine learning methods\cite{sheridan2013time}. As a result, we deploy two train-test splits that, compared to random splitting, are at once more challenging and also more accurately reflect the real world task of drug discovery. First, we split all datasets temporally, training on molecules assayed before the earliest $date_i$, selecting models based on performance on molecules assayed between $date_i$ and intermediate $date_j$, and evaluating the final model on held-out molecules assayed after the latest $date_j$. Such temporal splitting is meant to parallel the types of progressions that typically occur in lead optimization programs as well as reflect broader changes in the medicinal chemistry field.

\begin{figure*}[!htb]
\caption{Temporal plus Molecular Weight Split. In this rigorous cross-validation procedure, one trains and selects models based on older, smaller molecules, with the resultant best model (according to the validation set score) being evaluated a single time on a held out test set on newer, larger molecules. Specifically, the train and validation sets are comprised only of molecules below a certain molecular weight threshold and synthesized before a certain date, whereas the test set is comprised only of molecules above a certain molecular weight threshold and synthesized after a certain date.}
\label{cv_illustration}
\includegraphics[width=6in]{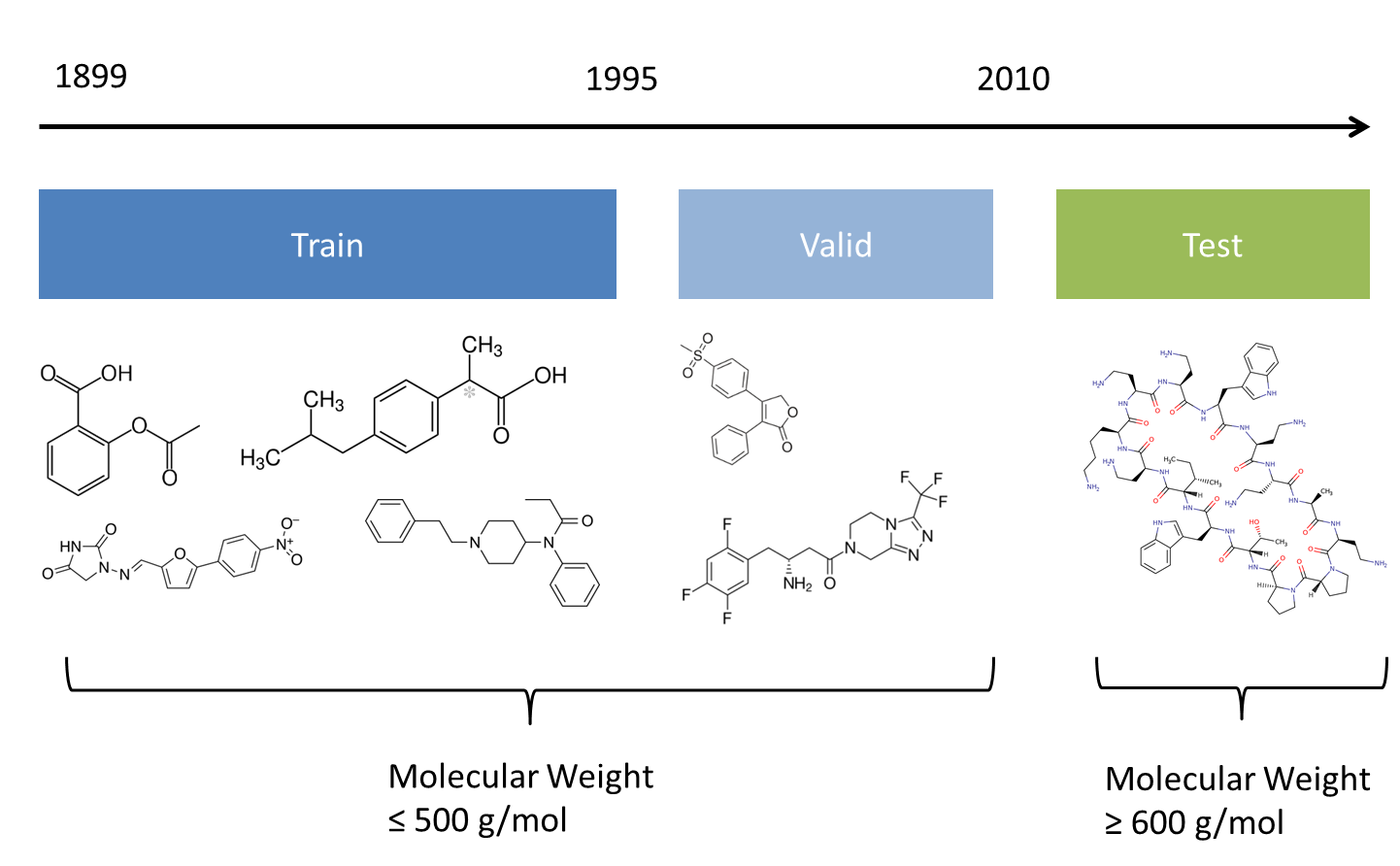}
\end{figure*}

In addition to temporal splitting, we introduce an additional cross-validation strategy in which we \textit{both} divide train, valid, and test sets temporally \textit{and} add the following challenge: (1) removal of molecules with molecular weight greater than $500 \frac{g}{mol}$ from the training and validation sets and (2) inclusion of only molecules with molecular weight greater than or equal to $600 \frac{g}{mol}$ from the test set. We denote this as \textit{temporal plus molecular weight} split (Figure \ref{cv_illustration}).

\subsection{Temporal Split}

\begin{figure*}[!htb]
\caption{Temporal Split: Performance of PotentialNet versus Random Forest for All Assays}
\label{cv1_barplot}
\includegraphics[width=7in]{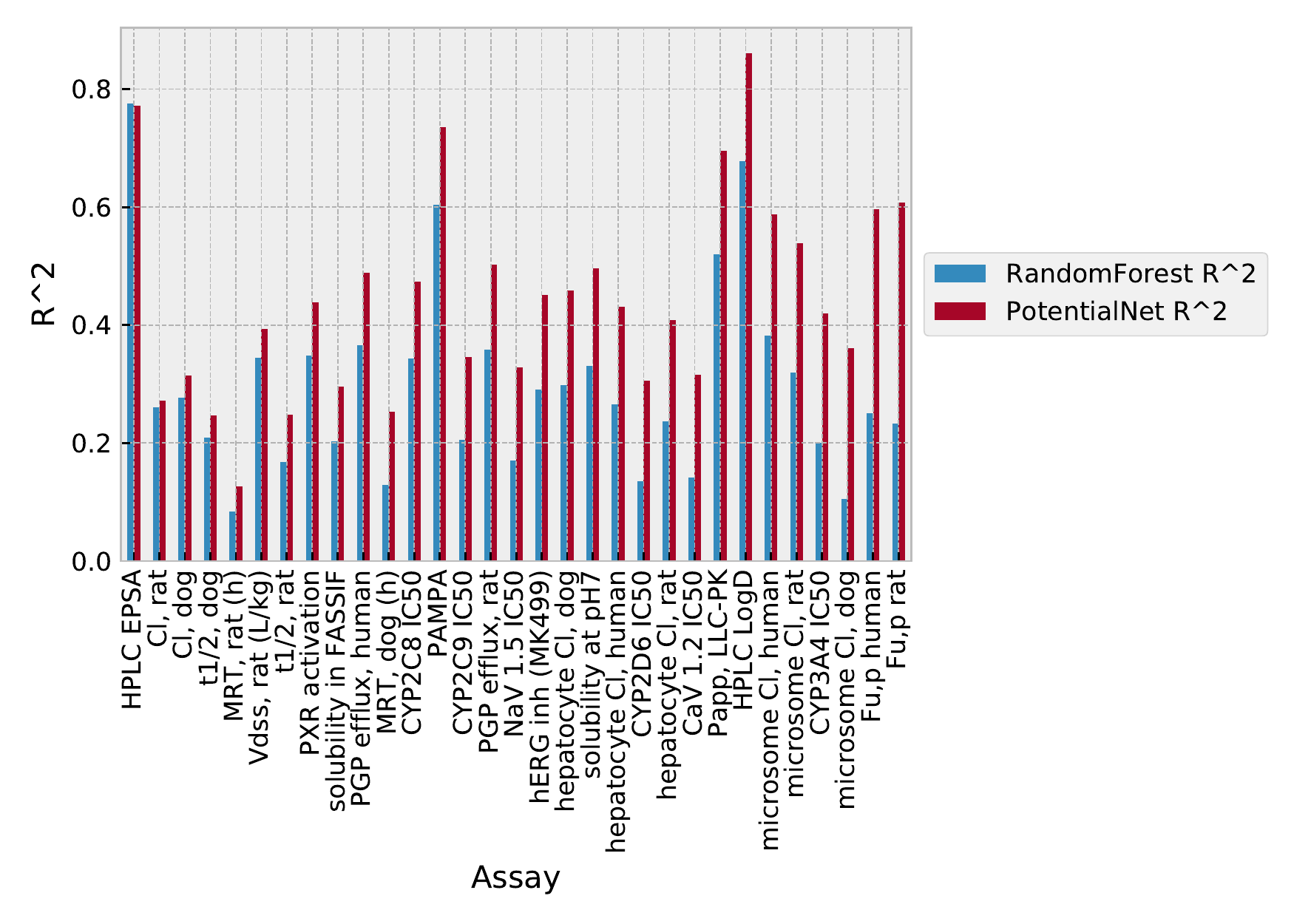}
\end{figure*}

\begin{figure*}[!htb]
\caption{Temporal Split: Scatter plots of predictions by PotentialNet and by Random Forest vs. Experiment for Several Assays}
\label{temporal_scattermatrix}
\includegraphics[width=7in]{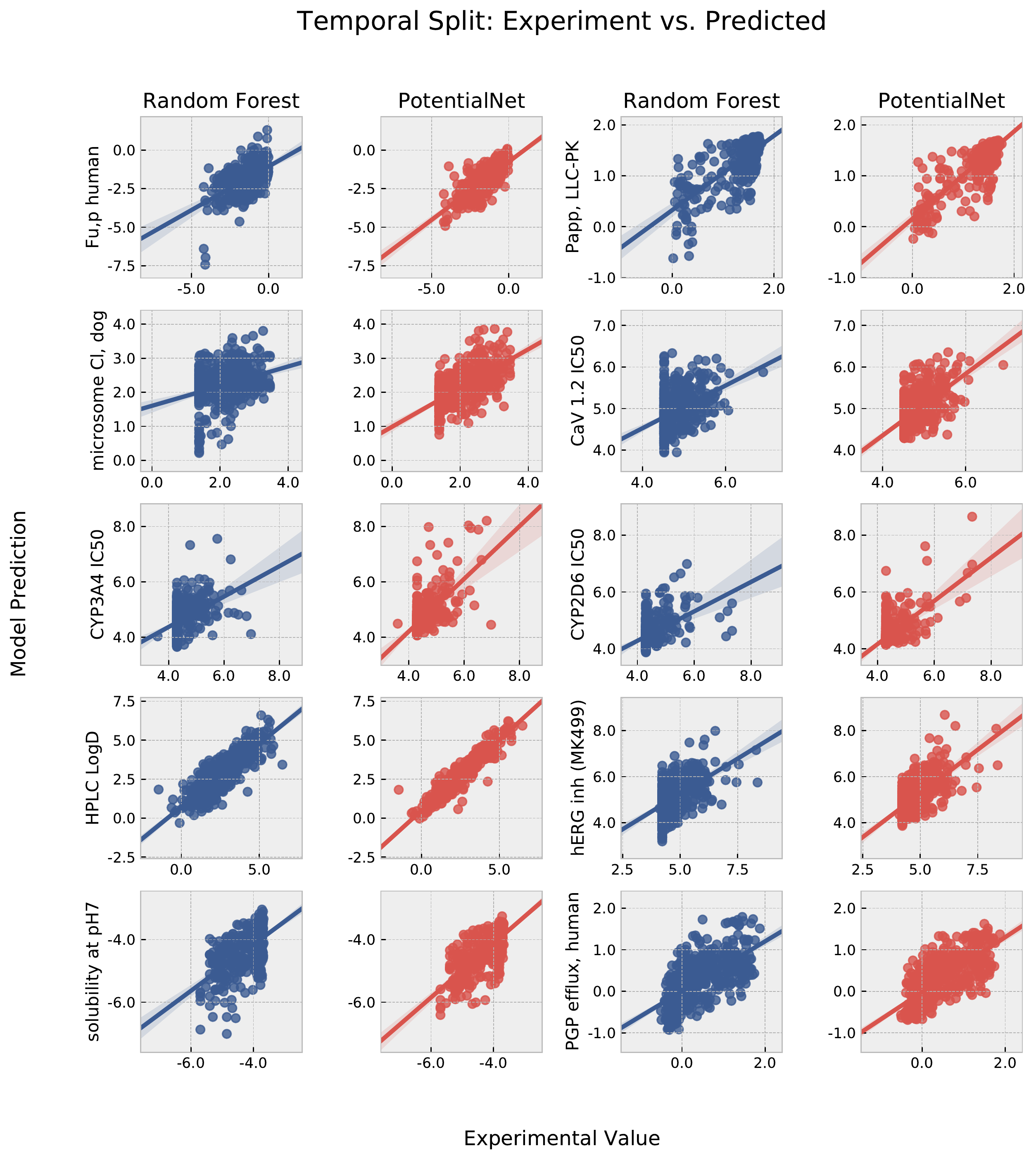}
\end{figure*}

In aggregate, PotentialNet achieves a $64\%$ average improvement and a $52\%$ median improvement in $R^2$ over Random Forests across all thirty-one reported datasets (Figure \ref{cv1_barplot}, Table \ref{temporal_agg}, Table \ref{temporal}). The mean $R^2$ over the various test datasets is $0.30$ for Random Forests and $0.44$ for PotentialNet, corresponding to a mean $\overline{\Delta R^2 } \left (PotentialNet - Random Forest \right) = 0.15$. Among the assays for which PotentialNet offers the most improvement (Figure \ref{temporal_scattermatrix}) are plasma protein binding (fraction unbound for both human, $\Delta R^2 = 0.34$ and rat, $\Delta R^2 = 0.38$), microsomal clearance (human: $\Delta R^2 = 0.21$, dog: $\Delta R^2 = 0.26$, rat: $\Delta R^2 = 0.22$), CYP3A4 Inhibition ($\Delta R^2 = 0.22$), logD ($\Delta R^2 = 0.18$), and passive membrane absorption ($\Delta R^2 = 0.18$). Meanwhile, assays HPLC EPSA and rat clearance, which account for two of the thirty-one assays, show no statistically significant difference. All $R^2$ values are reported with confidence intervals computed according to ref \cite{Walters2018}.

In addition to improvements in $R^2$, for both cross-validation schemes discussed, we note that the slope of the linear regression line between predicted and experimental data is, on average, closer to unity for PotentialNet than it is for Random Forests. This difference is qualitatively notable in Figure \ref{temporal_scattermatrix}. A corollary, which is also illustrated by Figures \ref{temporal_scattermatrix}, \ref{temporal_mw_scattermatrix}, and \ref{prospective_scattermatrix}, is that PotentialNet DNN's perform noticeably better than Random Forest in predicting the correct \textit{range} of values for a given prediction task. At our institution, this deficiency of RF is in part rectified by \textit{ex post facto} prediction rescaling, which in part recovers the slope but makes no difference in $R^2$.

The commercially available molecules for which PotentialNet achieved the greatest improvement in prediction versus Random Forests are displayed in Table \ref{small_cv}. An example of a molecule on which Random Forest renders a more accurate prediction is shown in Table \ref{small_cv_bad}.

\subsection{Temporal plus Molecular Weight Split}

\begin{figure*}[!htb]
\caption{Temporal plus Molecular Weight Split: Performance of PotentialNet versus Random Forest for All Assays}
\label{cv_mw50_barplot}
\includegraphics[width=7in]{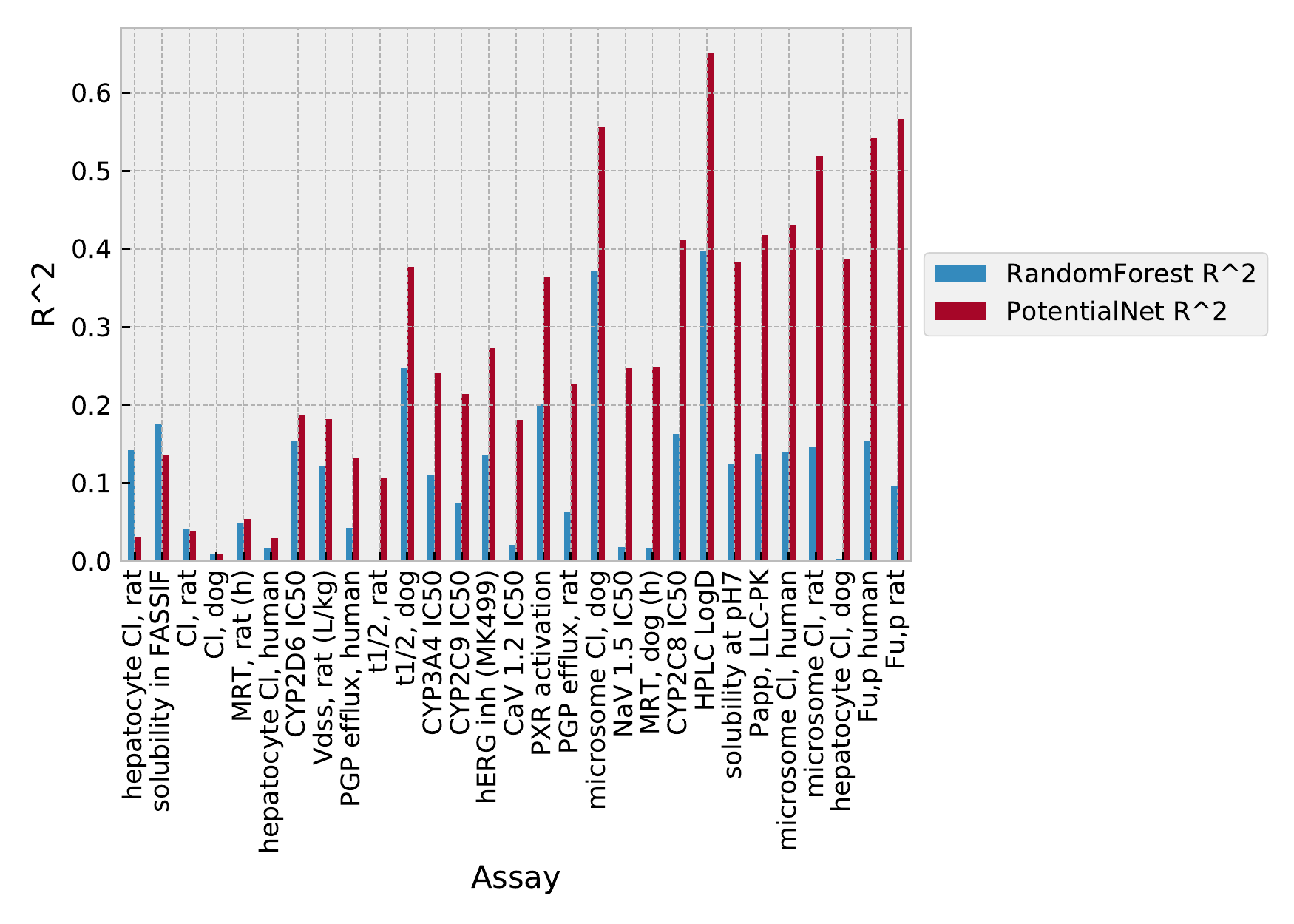}
\end{figure*}

\begin{figure*}[!htb]
\caption{Temporal plus Molecular Weight Split: Scatter plots of predictions by PotentialNet and by Random Forest vs. Experiment for Several Assays}
\label{temporal_mw_scattermatrix}
\includegraphics[width=7in]{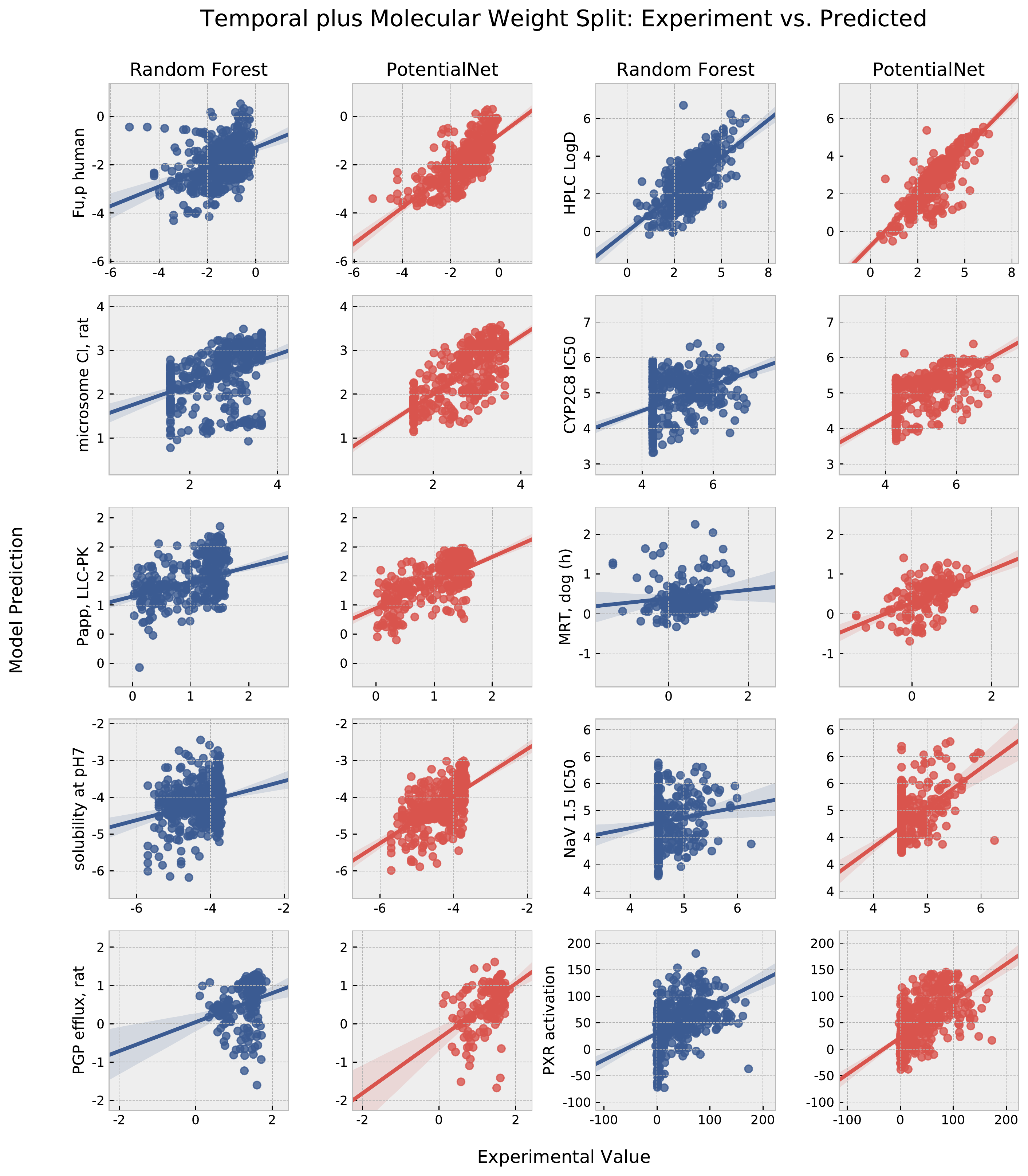}
\end{figure*}

We then investigated the cross-validation setting where data was both (a) split temporally and (b) molecules with $MW > 500 \frac{g}{mol}$ were removed from the training set while \textit{only} molecules with $MW > 600 \frac{g}{mol}$ were retained in the test set (Figure \ref{cv_illustration}). In aggregate, PotentialNet achieves a $153\%$ median improvement in $R^2$ over Random Forests across all twenty-nine reported datasets\footnote{Note that PAMPA and EPSA are not included due to insufficient number of compounds meeting the training and testing criteria.} (Table \ref{aggregate_mw500}, Figure \ref{cv_mw50_barplot}). The mean $R^2$ over the various test datasets is $0.12$ for Random Forests and $0.28$ for PotentialNet, corresponding to a mean $\Delta R^2 \left (PotentialNet - Random Forest \right) = 0.16$. The assays for which PotentialNet offers the most improvement (Figure \ref{temporal_mw_scattermatrix}) are plasma protein binding (fraction unbound for both human, $\Delta R^2 = 0.38$ and rat, $\Delta R^2 = 0.47$), microsomal clearance (human: $\Delta R^2 = 0.29$, dog: $\Delta R^2 = 0.19$, rat: $\Delta R^2 = 0.37$), CYP2C8 Inhibition ($\Delta R^2 = 0.25$), logD ($\Delta R^2 = 0.25$), and passive membrane absorption ($\Delta R^2 = 0.28$). Meanwhile, human hepatocyte clearance,  CYP2D6 Inhibition, rat and dog clearance, dog halflife, human PGP (efflux), rat MRT, and rat volume of distribution exhibit no statistically significant difference in model predictivity ($\frac{8}{29}$ of all datasets), with only rat hepatocyte clearance being predicted less well for PotentialNet as compared to Random Forests. It should be noted that the quantity of molecules in the test sets are smaller in temporal plus MW split as compared to temporal only split, and therefore, it is accordingly more difficult to reach statistically significant differences in $R^2$ (S.I. Table \ref{temporal_mw_nmols}).

The commercially available molecules for which PotentialNet achieved the greatest improvement in prediction versus Random Forests are displayed in Table \ref{big_cv}. It is intriguing that the same molecule undergoes the greatest improvement for both Human Fraction Unbound as well as CYP2D6 Inhibition.

As previous works have noted~\cite{xu2017demystifying}, multitask style training (cf. Methods) can boost -- or, less charitably, inflate -- the performance of neural networks by sharing information between the training molecules of one task and the test molecules of another task, especially if the activities are in some way correlated. Another advantage of the Temporal plus Molecular Weight cross-validation approach is that it mitigates hemorrhaging of information between assay datasets. By introducing a minimum $100 \frac{g}{mol}$ molecular weight gap between train and test molecules, it is not only impossible for train molecules in one task to appear as test molecules for another task, but it also circumscribes the similarity of any given task's training set to any given other task's test set. We further investigate the relative effect of multitask versus single task training in Supplementary Table \ref{st_vs_mt}. However, even in cases where there is similarity or even identity between training molecules of one assay and test molecules of another assay, in the practice of chemical machine learning, this may in fact be desirable in cases. For instance, if less expensive, cell-free assays, like solubility, have a strong correlation with a more expensive endpoint, like dog mean residence time, it would be an attractive property of multitask learning if solubility data on molecules in a preexisting database could inform more accurate predictions of the animal mean residence time of untested, similar molecules.

\subsection{Held out data from literature}

To further ascertain the generalization capacity of our models, we obtained data from scholarly literature. In particular, we obtained data on macrocyclic compounds for passive membrane permeability and logD from ref \cite{over2016structural}. We observed a statistically significant increase in performance (Table \ref{nchembio_r2}) for both passive membrane permeability ($\Delta R^2 \left (PotentialNet - Random Forest \right ) = 0.23$) and logD ($\Delta R^2 \left (PotentialNet - Random Forest \right ) = 0.21$). The four molecules for which PotentialNet exhibits the greatest improvement in predictive accuracy over Random Forests are shown in Table \ref{nchembio_mols}. 

The second molecule in Table \ref{nchembio_mols} is experimentally quite permeable, which PotentialNet correctly identifies but Random Forests severely underestimates. Note that the aliphatic tertiary amine would likely be protonated and therefore charged at physiologic pH. The proximity of an ether oxygen may ``protect'' the charge, increasing the ability to passively diffuse through lipid bilayers. Because of the relative efficiency with which information traverses bonds in a graph convolution as opposed to the fixed pair features that are provided to the random forest, it is intuitively straightforward for a graph neural network to learn the ``atom type'' of a high pKa nitrogen in spatial proximity to an electron rich oxygen, whereas pair features would rigidly specify an aliphatic nitrogen three bonds away from an aliphatic oxygen. 

\subsection{Feature Interpretation}

Feature interpretation remains a fledgling discipline in many areas of deep learning. We posit a simple method here to probe a given graph convolutional neural network to expose intuitive reasons driving that network's decision-making process. Recall that the basic graph convolutional update is $h^{(1)} = ReLU \left (W^{(1)} \cdot A \cdot X \right )$. For a given molecule $\hat{x}$ with predicted property $\hat{y}$, we can use the backpropagation algorithm to determine the gradient, or partial derivative per feature, on the input. We define the feature-wise importance $Imp(atom_i)$ of $atom_i$ as:

\begin{align}
\label{imp}
Imp(atom_i) = \sum_{j=0}^{(f_{in})} \frac{\partial NN(A,X)}{\partial X_{ij}}
\end{align}

where $f_{in}$ is the initial number of features per atom (number of columns of $X$) and $X_{ij}$ is the $j'th$ feature of $atom_i$. 

By a related metric, we can posit the substructure, or functional group, of size $S$ atoms of a molecule that has the greatest impact on the graph convolutional neural network's prediction by:

\begin{align}
\label{imp_subgraph}
argmax_{\text{subgraph G'}} \sum_{atom_j \in G'} Imp(atom_j)
\end{align}

\begin{figure*}[!htb]
\centering
\subfloat[Individually Most Important Atoms]{%
  \includegraphics[width=3in]{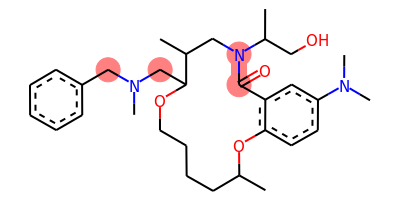}
  \label{nchembio_fourmost}%
}\qquad
\subfloat[Most Important Substructure]{%
  \includegraphics[width=3in]{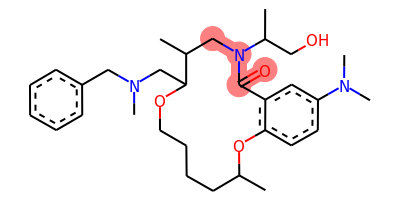}
  \label{nchembio_substruct}%
}
\caption{Feature interpretation for macrocycle in ref \cite{over2016structural}}
\end{figure*}

The above comprises a plausible route for feature interpretation in graph convolutional neural networks. While a rigorous evaluation of this approach will remain the subject of future work, we illustrate how it would function with a large molecule example. Let us reexamine the case of the molecule in Table \ref{nchembio_mols} that is correctly identified by PotentialNet as membrane permeable (and misidentified by Random Forests as impermeable). Intriguingly, the feature importance score (Equation \ref{imp}) points to the two carbons neighboring the tertiary amine nitrogen, and the amide carbon and nitrogen as the four most important atoms for determining permeability (Figure \ref{nchembio_fourmost}, Equation \ref{imp}). The importance of the tertiary amine adjacent atoms certainly correlates with chemical intuition. Meanwhile, the amide group is seen as important by both individual per-atom gradient ranking as well as by maximal substructure importance (Figure \ref{nchembio_substruct}, Equation \ref{imp_subgraph}). The interpretation of the amide's high importance is less obvious, though several studies\cite{rezai2006testing, hickey2016passive} have examined the influence of macrocycle amides and permeability. For instance, it has been proposed\cite{rezai2006testing} that amide groups in the main ring of macrocycles stabilize conformations that enable intramolecular hydrogen bond formation, thereby reducing the effective polar surface area of the molecule. It is possible that the graph neural network is learning a correlation between macrocyclic amides and reduced polar surface area.

\subsection{Prospective Study}

\subsubsection{Overall Prospective Performance on Nov, 2018 - Feb, 2019 Data}
\begin{figure*}[!htb]
\caption{Prospective Study. All model parameters for both PotentialNet and Random Forests were frozen in August, 2018. Subsequently, performance of both models was compared for new assay data collected on compounds registered from Nov, 2018 - Feb, 2019.}
\label{prospective_barplot}
\includegraphics[width=7in]{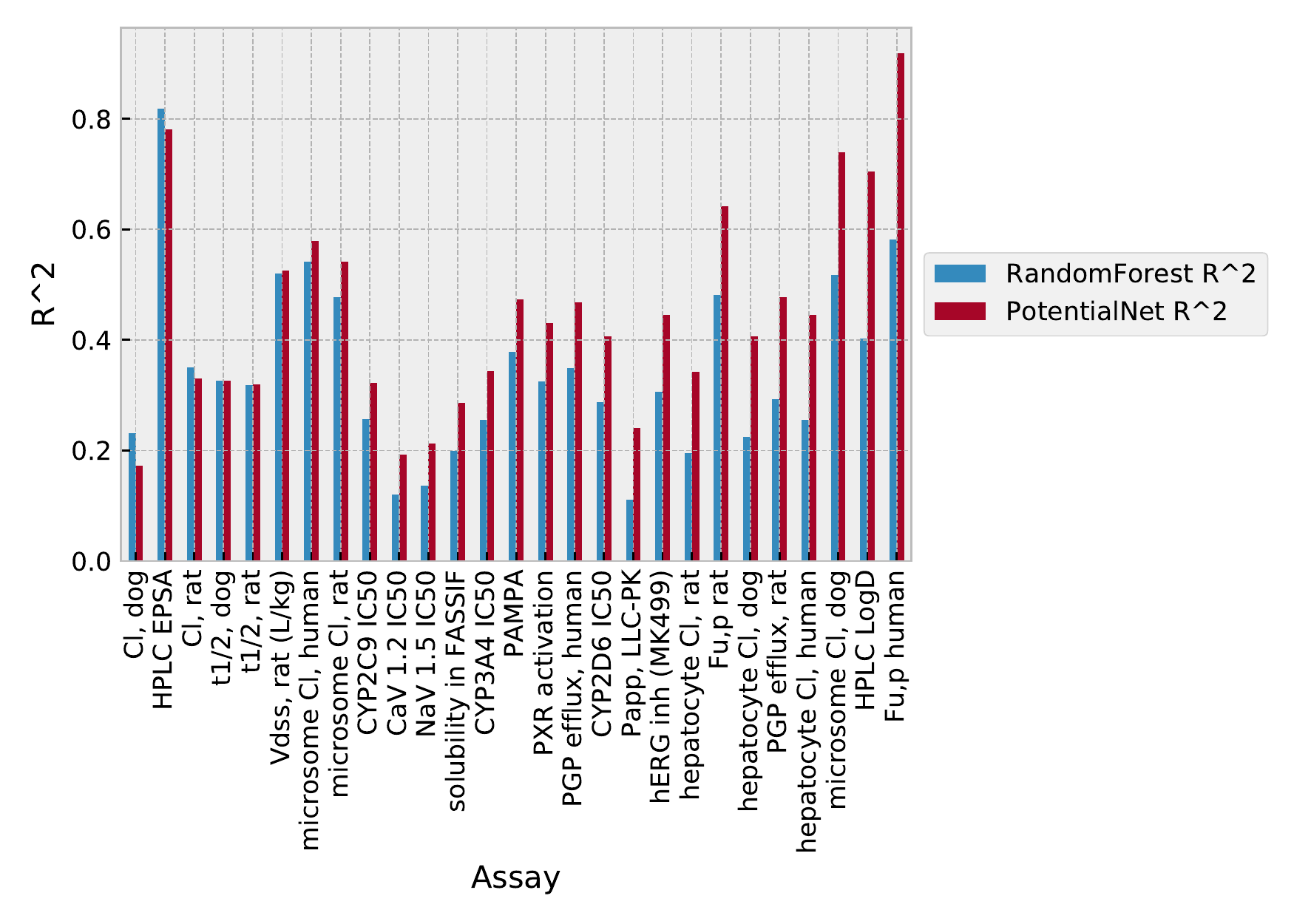}
\end{figure*}

\begin{figure*}[!htb]
\caption{Nov, 2018 - Feb, 2019 Prospective Data: Scatter plots of predictions by PotentialNet and by Random Forest vs. Experiment for Several Assays. All models were trained on data and compounds registered up through August, 2018 and tested prospectively on data and compounds registered in Nov, 2018 - Feb, 2019}
\label{prospective_scattermatrix}
\includegraphics[width=7in]{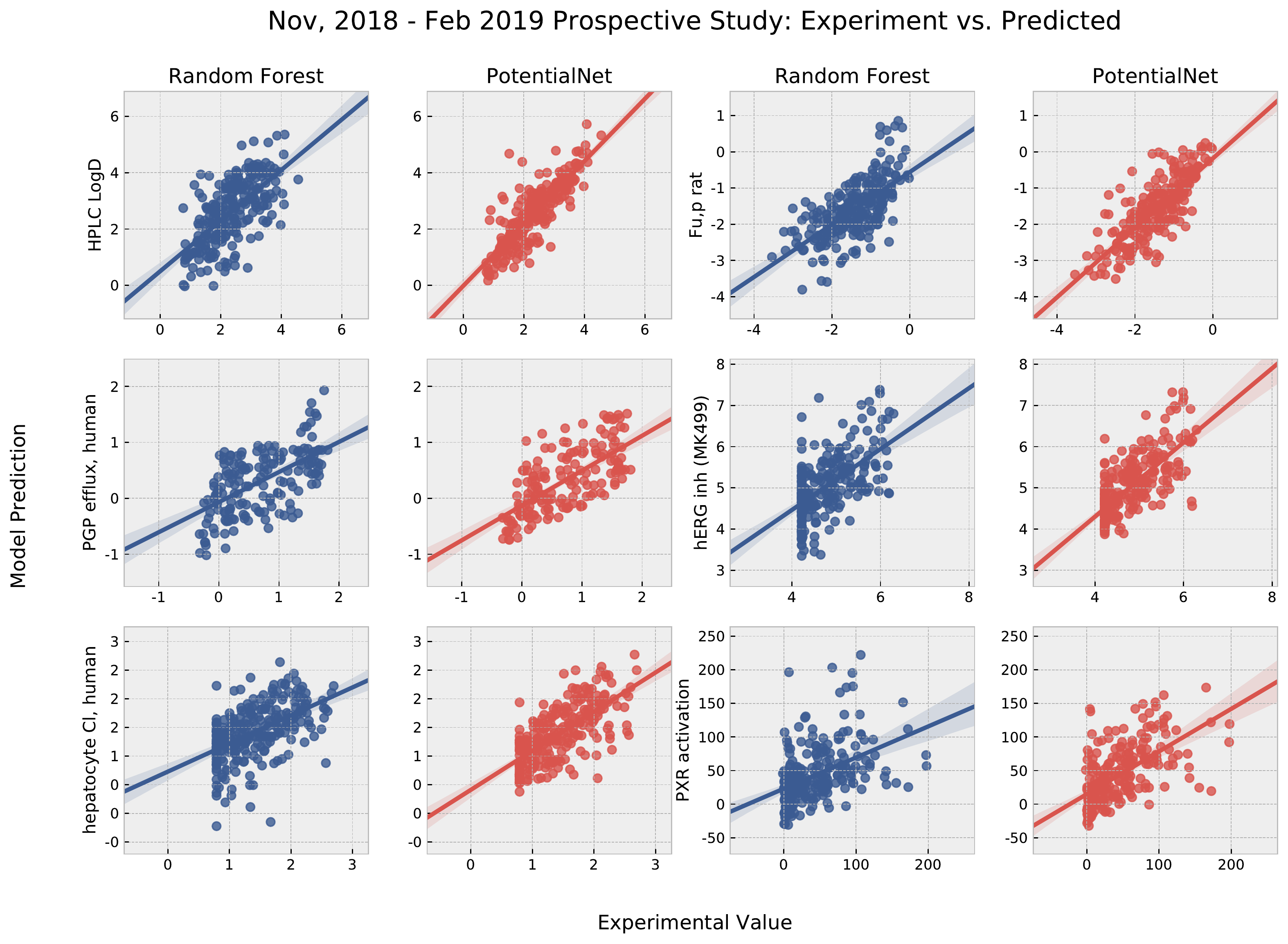}
\end{figure*}

In September, 2018, we froze the parameters of Random Forest and PotentialNet models trained on all available assay data recorded internally at Merck up to the end of August, 2018. After approximately two months had elapsed after the registration of the last training data point, we evaluated the performance of those \textit{a priori} frozen models on new experimental assay data gathered on compounds registered between November, 2018 and the end of February, 2019. Not only does this constitute a prospective analysis, but a particularly rigorous one in which there is a two month gap between training data collection and prospective model evaluation, further challenging the generalization capacity of the trained models. For statistical power, we chose to evaluate performance on all assays for which at least ten compounds were experimentally tested the period Nov, 2018 - Feb, 2019. Over these twenty-seven assays, Random Forest achieved a median $R^2$ of 0.32, whereas PotentialNet achieved a median $R^2$ of 0.43 for a median $\Delta R^2 = 0.10$ (Table \ref{prospective_agg}). Performance of each assay can be found in Figure \ref{prospective_barplot} and Table \ref{prospective}, and scatter plots of predicted versus experimental values for several assays can be found in Figure \ref{prospective_scattermatrix}. While it makes no difference in $R^2$, we have chosen to scale the values predicted by both Random Forests and by PotentialNet to match the mean and standard deviation of the distribution of assay data in the training set to more faithfully reflect how these models would be used practically in an active pharmaceutical project setting.

\subsubsection{Performance on Two Specific Projects}

\begin{figure*}[!htb]
\caption{Prospective prediction of rat fraction unbound in plasma (rat fu,p) in two active projects using random forest models (top row) and potential net models (bottom row), for compounds experimentally tested between September and December 2018.  Data for Project 1 and Project 2 is for 97 and 123 compounds, respectively.}
\label{prospective_fu}
\includegraphics[width=7in]{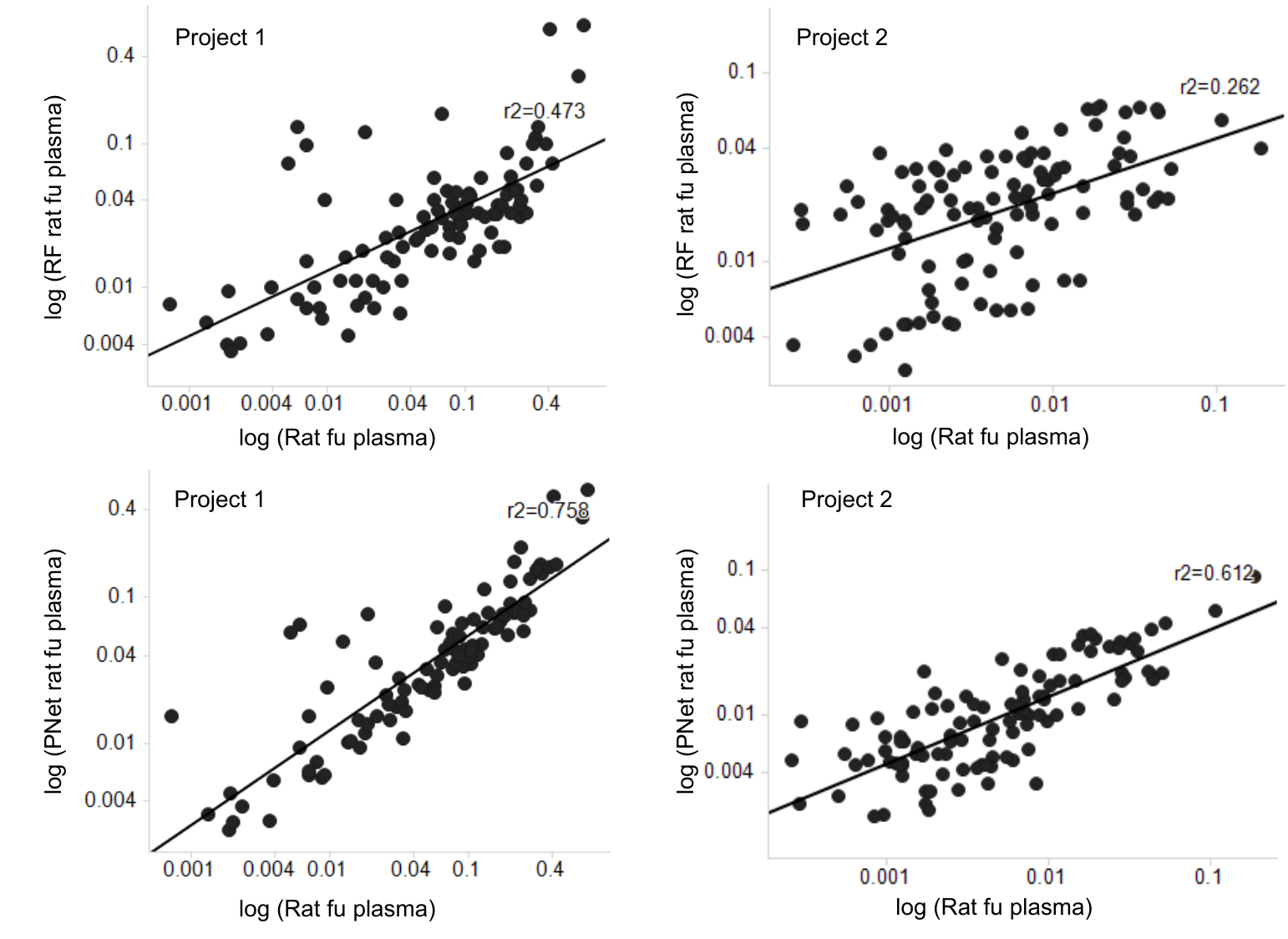}
\end{figure*}

To assess performance on individual projects, we applied the August, 2018, models to prediction of rat plasma fraction unbound on two currently active lead optimization projects at Merck.  The results (Figure \ref{prospective_fu}) suggest that the performance on the individual projects are similar to the temporal split and temporal + molecular weight split results.

\section{Discussion}
Preclinical drug discovery is a critica, and often rate-limiting stage of the broader pharmaceutical development pipeline. While estimates vary between studies, recent analyses estimate the capitalized cost of preclinical discovery per FDA-approved drug as anywhere between \$89 Million and \$834 Million \cite{dimasi2003price, morgan2011cost, paul2010improve}. The multi-objective optimization among potency and ADMET properties, which can entail vexing trade-offs, is a critical bottleneck in preclinical discovery \cite{wager2010defining, segall2011applying}. More accurate prediction of ADMET endpoints can both prevent exploration of undesirable chemical space as well as facilitate access to desirable regions of chemical space, thereby making preclinical discovery not only more efficient but perhaps more productive as well. 

To assess if a modern graph convolutional neural network \cite{feinberg2018potentialnet} succeeds in more accurately predicting ADMET endpoints, we conducted a rigorous performance comparison between GCNN and the previous state-of-the-art random forest based on cheminformatic features. With an emphasis on rigor, we included a total of thirty-one assay datasets in our analysis, employed two cross-validation splits (\textit{temporal split} \cite{sheridan2013time} and a combined \textit{temporal plus molecular weight split}), and made predictions on a publicly available held-out test set. Finally, we made prospective predictions with both random forests and with PotentialNet and then compared with experimental results.

Encouragingly, statistical improvements were observed in each of the four aforementioned validation settings. In the temporal split setting, across thirty-one tasks, Random Forests achieved a mean $R^2$ of 0.30, whereas PotentialNet achieved a mean $R^2$ of 0.44 (Table \ref{temporal_agg}). In the temporal plus molecular weight split setting -- where only older smaller molecules were included in the training set while only newer larger molecules included in the test set -- across twenty-nine tasks, Random Forests achieved a mean $R^2$ of 0.12, whereas PotentialNet achieved a mean $R^2$ of 0.28 (Table \ref{aggregate_mw500}). In the final pseudo-prospective validation setting, we assessed the ability of pre-trained Random Forests and pre-trained PotentialNet models to predict passive membrane permeability and logD on an experimental dataset on macrocycles obtained from the literature\cite{over2016structural}. In this setting, for passive membrane permeability, Random Forests achieved an $R^2$ of 0.15 whereas PotentialNet achieved an $R^2$ of 0.38; for logD, Random Forests achieved an $R^2$ of 0.39 whereas PotentialNet achieved an $R^2$ of 0.60 (Table \ref{big_cv}).

While the three described retrospective investigations are more rigorous than random splitting and are meant to more faithfully reflect the generalization capacity of a model in the practical real world of pharmaceutical chemistry, we also believe that prospective validation is important whenever the resources are available to do so. To this end, we made predictions on twenty-three assays, each of which contained measurements for new chemical entities synthesized and evaluated after November, 2018 (the last data point for model training was collected in August, 2018). In aggregate, there is a mean  $\Delta R^2$ of $0.10$ of PotentialNet over Random Forests. This improvement in accuracy in a future and relatively constrained time window is largely consistent with that prognosticated by the retrospective temporal split study and is encouraging for the utility of deep featurization in a predictive capacity for drug discovery.

Historically, as a discipline, machine learning arose from statistical learning, and a key line of inquiry in statistics involves extricating potentially confounding variables. Compared with random forests, we introduce several algorithmic changes at once: use of neural network instead of random forest; use of graph convolution as a neural network architecture based on a graph adjacency and feature matrices as input rather than either RF or MLP based on flat 1D features; and use of a variant of multitask learning rather than single task learning. How much of the performance gain accrued by PotentialNet can be attributed to each of the aforementioned changes? To investigate, we conducted an algorithm ablation study to compare performance contributions (Supplementary Tables \ref{st_vs_mt} and \ref{st_vs_mt_temporal}; we also include xgboost for additional comparison). It is reasonable to contend that one should solely compare random forest with single task neural networks since the former is incapable of jointly learning on several assay datasets simultaneously. However, one of the intuitive advantages of a GCNN over \textit{either} RF or MLP is that a GCNN can learn the atomic interaction features relevant to the prediction task at hand. Therefore, we aver that the most reflective comparison is to apply best practices that are accessible by each technique. Not only can graph convolutions learn the features, but adding different molecules from different tasks allows networks to learn more accurately both through the effect of task correlation and learning richer features by incorporating a greater area of chemical space.

While we are restricted with respect to the compounds in our training data that we can disclose, we can share select publicly disclosed compounds that happened to have been tested in the assays discussed in this work. Table \ref{small_cv} lists commercially available compounds for which PotentialNet's predictions are most improved compared to Random Forests' predictions in the temporal split setting. For example, the first compound, Methyl 4-chloro-2-iodobenzoate, has an experimental logD of 3.88, Random Forests predicts logD to be 2.26, and PotentialNet predicts logD to be 3.70. Neural network interpretation remains a discipline in its infancy and therefore renders it challenging to pinpoint exactly which aspect of either the initial featurization or the network enables PotentialNet to properly estimate the logD while Random Forests significantly underestimates it. As a hypothetical analysis, pair features would include such terms as ``carbonyl oxygen that is three bonds away from ether carbon,'' ``carbonyl oxygen that is four bonds away from aromatic iodine,'' and ``carbonyl oxygen that is six bonds away from an aromatic chlorine.'' There is no sense that it is the \textit{same} carbonyl carbon that has all of these properties. In stark contrast, by recursively propagating information, a graph convolution would confer a single, dense ``atom type'' on the carbonyl oxygen that would reflect its identity as a halogenated benzaldehyde oxygen. 

More accurate prediction of ADMET endpoints can be a torch\cite{paszke2017automatic}light guiding creative medicinal chemists as they explore uncharted chemical space en route to the optimal molecule. The results delineated in this paper demonstrate that deep-feature learning with graph convolutions can systematically and often quite significantly outperform random forests based on fixed fingerprints. We therefore deem it advisable for pharmaceutical scientists to consider integrating deep learning in general and graph convolutions in particular in their modeling pipelines.

\section*{Methods}

\begin{figure*}[!htb]%[H]
\caption{Multitask framework for PotentialNet training}
\label{multitask_framework}
\includegraphics[width=7in]{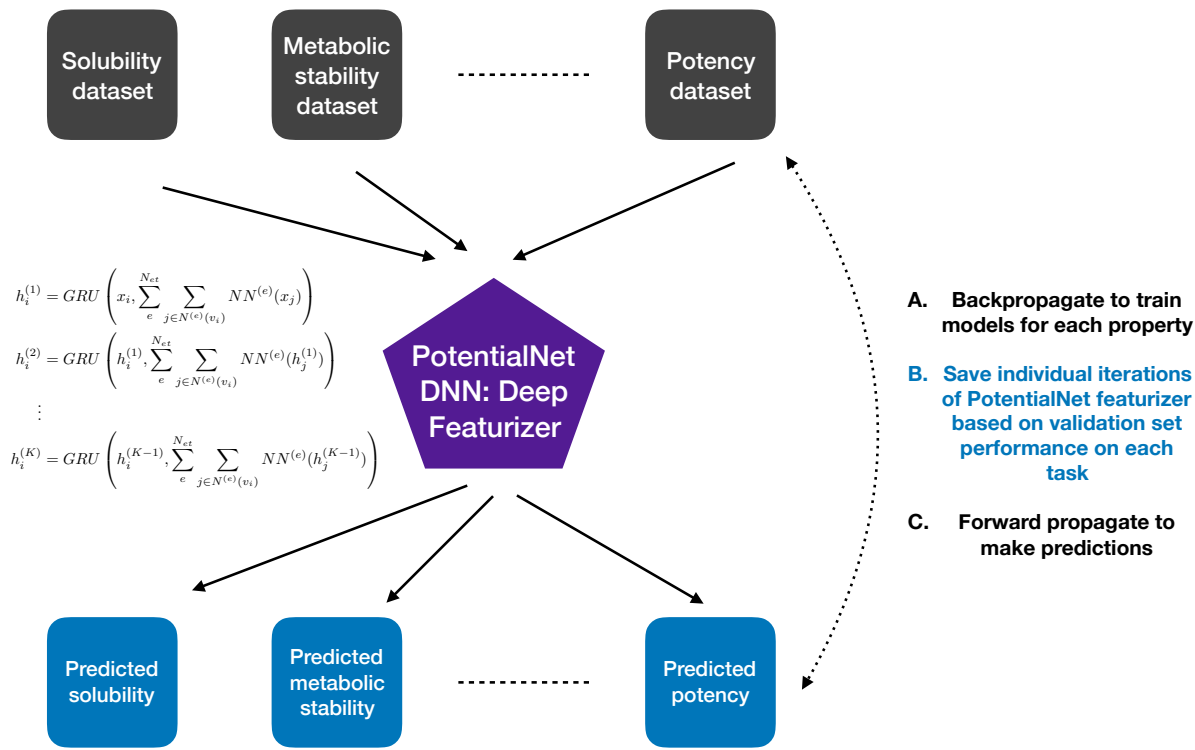}
\end{figure*}

PotentialNet (Equation \ref{S-GGNN}) \cite{feinberg2018potentialnet} neural networks were constructed and trained with PyTorch~\cite{paszke2017automatic}. Multilayer perceptron (MLP) neural networks were trained with the assistance of the MIX library that is internal to Merck (more details below). Following previous works \cite{ramsundar2015massively}, we make extensive use of multitask learning to train our PotentialNet models. We modified the standard multitask framework to save different models for each task on the epoch at which performance was best for that specific task on the validation set (Figure \ref{multitask_framework}). In that way, we employ an approach that draws on elements of both single and multitask learning. Custom Python code was used based on RDKit~\cite{rdkit} and OEChem~\cite{oechem} with frequent use of NumPy~\cite{numpy} and SciPy~\cite{scipy}. Networks were trained on chemical element, formal charge, hybridization, aromaticity, and the total numbers of bonds, hydrogens (total and implicit), and radical electrons. Random forest were implemented using both scikit-learn~\cite{sklearn} and MIX; all sklearn-trained random forests models were trained with $500$ trees and $\sqrt{n_{features}}$ per tree; xgboost models were trained using MIX.

QSAR Descriptors: Chemical descriptors, termed ``APDP'' used in this study for random forests, xgboost, and MLP DNN's are listed as follows. All descriptors are used in frequency form, i.e. we use the number of occurrences in a molecule and not just the binary presence or absence. APDP denotes the union of AP, the original ``atom pair'' descriptor from ref \cite{carhart1985atom}, and DP descriptors (``Donor acceptor Pair''), called ``BP'' in ref. \cite{kearsley1996chemical}. Such APDP descriptors are used in most of Merck's QSAR studies and in Merck's production QSAR. Both descriptors are of the form: ``Atom $type_i$ – (distance in bonds) – Atom $type_j$''

For AP, atom type includes the element, number of nonhydrogen neighbors, and number of pi electrons; it is very specific. For DP, atom type is one of seven (cation, anion, neutral donor, neutral acceptor, polar, hydrophobe, and other); it contains a more generic description of chemistry.

QSAR methods: All methods are used in regression mode, i.e. both input activities and predictions are floating-point numbers. All appropriate descriptors are used in the models, i.e. no feature selection is done. When random forests are not trained with scikit-learn, they are trained with the Merck MIX library that in turn  calls the R module RandomForest \cite{breiman2011package}, which encodes the original method of ref. \cite{breiman2001random} and first applied to QSAR in ref.\cite{svetnik2003random}. The default settings are 100 trees, nodesize=5, mtry=M/3 where M is the number of unique descriptors.

MLP Deep neural networks (DNN): We use Python-based code obtained from the Kaggle contest and described in ref \cite{ma2015deep}. We use parameters slightly different than the “standard set” described in that paper: Two intermediate layers of 1000 and 500 neurons with 25\% dropout rate and 75 training epochs. The above change is made for the purposes of more time-efficient calculation. The accuracy of prediction is very similar to that of the standard set.

xgboost: Extreme Gradient Boosting method published in ref \cite{chen2016xgboost}. In this paper, we are using a set of standard parameters from Merck's subsequent study using this method \cite{sheridan2016extreme}.

\section{Tables}

\begin{table*}[ht]
	\caption{Aggregate performance: Temporal Split}
	\label{temporal_agg}
\begin{tabular}{lr}
\toprule
{} Metric &        Value \\
\midrule
Mean RandomForest R\textasciicircum2             &   0.30 \\
Mean PotentialNet R\textasciicircum2             &   0.44 \\
Median RandomForest R\textasciicircum2           &   0.27 \\
Median PotentialNet R\textasciicircum2           &   0.43 \\
Mean Absolute R\textasciicircum2 Improvement     &   0.15 \\
Mean Percentage R\textasciicircum2 Improvement   &   64\% \\
Median Absolute R\textasciicircum2 Improvement   &   0.16 \\
Median Percentage R\textasciicircum2 Improvement &   52\% \\
\bottomrule
\end{tabular}
\end{table*}

\begin{table*}
\caption{Aggregate Statistics: Temporal plus Molecular Weight Split}
\begin{tabular}{lr}
\toprule
{} Metric &        Value \\
\midrule
Mean RandomForest R\textasciicircum2             &     0.12 \\
Mean PotentialNet R\textasciicircum2             &     0.28 \\
Median RandomForest R\textasciicircum2           &     0.12 \\
Median PotentialNet R\textasciicircum2           &     0.25 \\
Mean Absolute R\textasciicircum2 Improvement     &     0.16 \\
Mean Percentage R\textasciicircum2 Improvement   &  1477\% \\
Median Absolute R\textasciicircum2 Improvement   &     0.16 \\
Median Percentage R\textasciicircum2 Improvement &   153\% \\
\bottomrule
\end{tabular}
\label{aggregate_mw500}
\end{table*}

\begin{table*}[ht]
	\caption{Aggregate performance: Prospective}
	\label{prospective_agg}
\begin{tabular}{lr}
\toprule
{} Metric &        Value \\
\midrule
Mean RandomForest R\textasciicircum2             &   0.34 \\
Mean PotentialNet R\textasciicircum2             &   0.45 \\
Median RandomForest R\textasciicircum2           &   0.32 \\
Median PotentialNet R\textasciicircum2           &   0.43 \\
Mean Absolute R\textasciicircum2 Improvement     &   0.10 \\
Mean Percentage R\textasciicircum2 Improvement   &   37\% \\
Median Absolute R\textasciicircum2 Improvement   &   0.10 \\
Median Percentage R\textasciicircum2 Improvement &   35\% \\
\bottomrule
\end{tabular}
\end{table*}

\begin{table*}[ht]
	\caption{Temporal Split: Commercially Available Molecules for which performance improvement of PotentialNet with graph features over Random Forest with pair features is greatest}
    \label{small_cv}
    \begin{tabular}{l l l l l l l}
    \hline
   Molecule & Property & Exp. & RandomForest  & PotentialNet & SMILES \\ \hline
		 \includegraphics[trim=0 0 0 -5,scale=0.3]{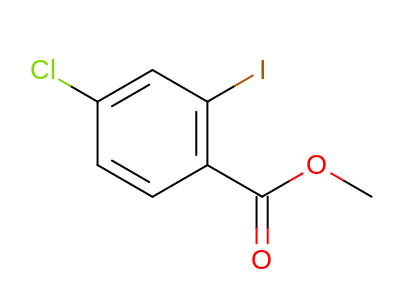} & logD & 3.88 & 2.26 & 3.70 & COC(=O)c1ccc(cc1I)Cl \\ \hline
		 \includegraphics[trim=0 0 0 -5,scale=0.3]{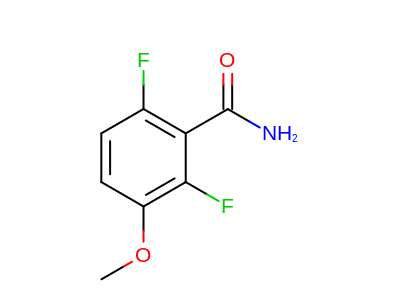} & logD & -0.61  & 1.58 & 0.59 & COc1ccc(c(c1F)C(=O)N)F \\ \hline
		 \includegraphics[trim=0 0 0 -5,scale=0.3]{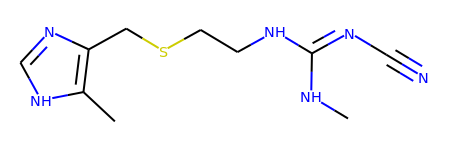} & Human Fraction Unbound & -0.06 & -1.30 & -0.45 & \shortstack{Cc1c(nc[nH]1) \\ CSCCN/C(=N/C\#N)/NC} \\ \hline
		 \includegraphics[trim=0 0 0 -5,scale=0.3]{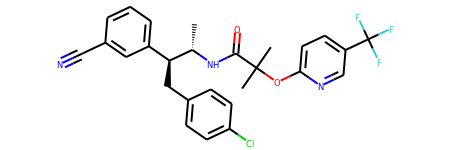} & Human Fraction Unbound & -2.95 & -2.24 & -3.07 & \shortstack{\tiny{C[C@@H]([C@@H](Cc1ccc(cc1)Cl)} \\ \tiny{c2cccc(c2)C\#N)NC(=O)C(C)} \\ \tiny{(C)Oc3ccc(cn3)C(F)(F)F}} \\ \hline
		 \includegraphics[trim=0 0 0 -5,scale=0.3]{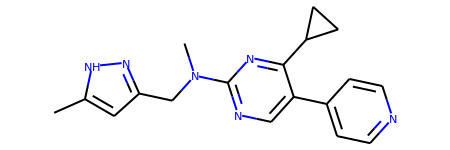} & CYP3A4 Inhibition & 6.14 & 4.61 & 5.54 & \shortstack{Cc1cc(n[nH]1)CN(C)c2ncc \\ (c(n2)C3CC3)c4ccncc4} \\ \hline
		 \includegraphics[trim=0 0 0 -5,scale=0.3]{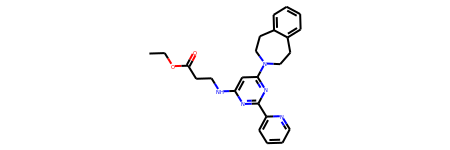} & hERG Inhibition & 6.62 & 5.17 & 5.80 & \shortstack{CCOC(=O)CCNc1cc(nc(n1)\\c2ccccn2)N3CCc4ccccc4CC3}
    \end{tabular}
\end{table*}

\begin{table*}[ht]
	\caption{Temporal Split: Example of molecule predicted more accurately by Random Forests than by PotentialNet}
	\label{small_cv_bad}
	\begin{tabular}{l l l l l l l}
	\hline
	   Molecule & Property & Exp. & RandomForest  & PotentialNet & SMILES \\ \hline
		\includegraphics[trim=0 0 0 -5,scale=0.3]{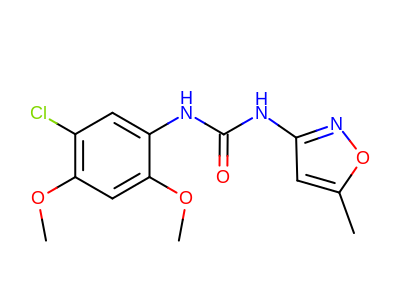} & Human Fraction Unbound & -1.51 & -1.55 & 1.86 & \shortstack{Cc1cc(no1)NC(=O)\\Nc2cc(c(cc2OC)OC)Cl} \\ \hline
	\end{tabular}
\end{table*}

\begin{table*}[ht]
	\caption{Performance: Temporal Split}
	\label{temporal}
\begin{tabular}{lrlrlrr}
\toprule
{} Dataset &  \shortstack{RandomForest \\ R\textasciicircum2} & \shortstack{RandomForest \\R\textasciicircum2, 95\% CI} &  \shortstack{PotentialNet \\ R\textasciicircum2} & \shortstack{PotentialNet \\R\textasciicircum2, 95\% CI} &  \shortstack{Absolute\\Improvement} & \shortstack{Percentage\\ Improvement} \\
\midrule
HPLC EPSA            &             0.775 &            (0.76, 0.789) &             0.772 &           (0.757, 0.786) &                -0.003 &                  -0.340 \\
Cl, rat              &             0.260 &            (0.25, 0.271) &             0.272 &           (0.262, 0.282) &                 0.011 &                   4.360 \\
Cl, dog              &             0.277 &           (0.258, 0.295) &             0.314 &           (0.296, 0.333) &                 0.038 &                  13.579 \\
t1/2, dog            &             0.209 &           (0.192, 0.225) &             0.247 &            (0.23, 0.264) &                 0.038 &                  18.455 \\
MRT, rat (h)         &             0.084 &           (0.075, 0.093) &             0.127 &           (0.117, 0.138) &                 0.044 &                  52.333 \\
Vdss, rat (L/kg)     &             0.345 &           (0.335, 0.356) &             0.393 &           (0.383, 0.403) &                 0.048 &                  13.769 \\
t1/2, rat            &             0.168 &            (0.16, 0.177) &             0.248 &           (0.239, 0.258) &                 0.080 &                  47.625 \\
\bottomrule
\end{tabular}
\end{table*}

\begin{table*}
	\caption{Performance: Temporal plus Molecular Weight Split}
	\label{cv_mw500}
\begin{tabular}{lrlrlrr}
\toprule
{} Dataset &  \shortstack{RandomForest \\ R\textasciicircum2} & \shortstack{RandomForest \\R\textasciicircum2, 95\% CI} &  \shortstack{PotentialNet \\ R\textasciicircum2} & \shortstack{PotentialNet \\R\textasciicircum2, 95\% CI} &  \shortstack{Absolute\\Improvement} & \shortstack{Percentage\\ Improvement} \\
\midrule
hepatocyte Cl, rat   &             0.142 &           (0.087, 0.206) &             0.030 &           (0.007, 0.068) &                -0.112 &                 -78.891 \\
solubility in FASSIF &             0.176 &           (0.158, 0.194) &             0.136 &            (0.12, 0.153) &                -0.040 &                 -22.804 \\
Cl, rat              &             0.041 &           (0.025, 0.061) &             0.039 &           (0.023, 0.059) &                -0.002 &                  -5.284 \\
Cl, dog              &             0.008 &             (0.0, 0.037) &             0.008 &             (0.0, 0.037) &                -0.000 &                  -0.281 \\
MRT, rat (h)         &             0.049 &           (0.028, 0.074) &             0.054 &            (0.032, 0.08) &                 0.005 &                  10.590 \\
hepatocyte Cl, human &             0.017 &           (0.003, 0.044) &             0.029 &           (0.008, 0.062) &                 0.012 &                  67.114 \\
CYP2D6 IC50          &             0.154 &           (0.134, 0.175) &             0.188 &            (0.166, 0.21) &                 0.034 &                  21.943 \\
\bottomrule
\end{tabular}
\end{table*}

\begin{table*}[ht]
	\caption{Performance: Prospective Split}
	\label{prospective}
\begin{tabular}{lrlrlrr}
\toprule
{} Dataset &  \shortstack{RandomForest \\ R\textasciicircum2} & \shortstack{RandomForest \\R\textasciicircum2, 95\% CI} &  \shortstack{PotentialNet \\ R\textasciicircum2} & \shortstack{PotentialNet \\R\textasciicircum2, 95\% CI} &  \shortstack{Absolute\\Improvement} & \shortstack{Percentage\\ Improvement} \\
\midrule 
Cl, dog              &             0.231 &           (0.163, 0.304) &             0.173 &           (0.111, 0.242) &                -0.058 &                 -25.162 \\
HPLC EPSA            &             0.819 &           (0.806, 0.831) &             0.781 &           (0.765, 0.795) &                -0.039 &                  -4.710 \\
Cl, rat              &             0.350 &           (0.317, 0.384) &             0.330 &           (0.297, 0.363) &                -0.020 &                  -5.770 \\
t1/2, dog            &             0.326 &           (0.257, 0.396) &             0.326 &           (0.257, 0.395) &                -0.001 &                  -0.162 \\
t1/2, rat            &             0.318 &           (0.285, 0.351) &             0.319 &           (0.286, 0.352) &                 0.001 &                   0.215 \\
Vdss, rat (L/kg)     &             0.520 &             (0.49, 0.55) &             0.525 &           (0.494, 0.554) &                 0.004 &                   0.846 \\
microsome Cl, human  &             0.542 &           (0.503, 0.579) &             0.579 &           (0.542, 0.614) &                 0.037 &                   6.878 \\
\bottomrule
\end{tabular}
\end{table*}

\begin{table*}[ht]
	\caption{Temporal plus Molecular Weight Split: Commercially Available Molecules for which performance improvement of PotentialNet with graph features over Random Forest with pair features is greatest}
	\label{big_cv}
	\begin{tabular}{l l l l l l l}
	\hline
    Molecule & Property & Exp. & RandomForest  & PotentialNet & SMILES \\ \hline
    	\includegraphics[trim=0 0 0 -5,scale=0.3]{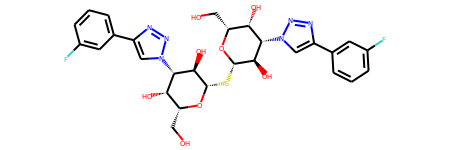} & Rat Fraction Unbound & -0.85 & -2.17 & -1.09 & \shortstack{c1cc(cc(c1)F)c2cn\\(nn2)[C@H]3[C@H]\\([C@H](O[C@H]\\([C@@H]3O)S[C@H]\\4[C@@H]([C@H]\\ ([C@H]([C@H](O4)CO)O)n5cc\\(nn5)c6cccc(c6)F)O)CO)O} \\ \hline
    	\includegraphics[trim=0 0 0 -5,scale=0.3]{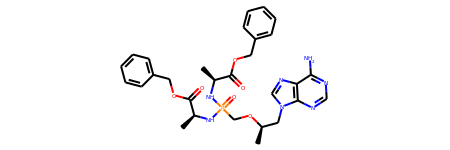} & Human Fraction Unbound & -1.17 & -2.15 & -1.60 & \shortstack{C[C@H](Cn1cnc2c1ncnc2N)\\OCP(=O)(N[C@@H]\\(C)C(=O)OCc3ccccc3)\\N[C@@H](C)C(=O)OCc4ccccc4} \\ \hline
    	\includegraphics[trim=0 0 0 -5,scale=0.3]{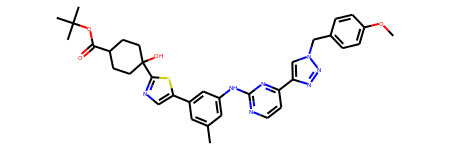} & logD & 4.55 & 3.02 & 4.43 & \shortstack{Cc1cc(cc(c1)\\Nc2nccc(n2)\\c3cn(nn3)Cc4ccc\\(cc4)OC)c5cnc(s5)C6\\(CCC(CC6)C(=O)OC(C)(C)C)O} \\ \hline
    	\includegraphics[trim=0 0 0 -5,scale=0.3]{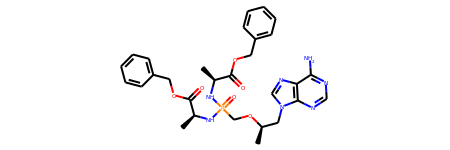} & CYP2D6 Inhibition & 4.30 & 4.95 & 4.53 & \shortstack{C[C@H](Cn1cnc2c1ncnc2N)\\OCP(=O)(N[C@@H](C)C\\(=O)OCc3ccccc3)N[C@@H]\\(C)C(=O)OCc4ccccc4} \\ \hline
	\end{tabular}
\end{table*}

\begin{table*}[ht]
	\caption{Performance on Held Out Data from Literature\cite{over2016structural}}
    \label{nchembio_r2}
    \begin{tabular}{l l l l}
    \hline
    Property & RandomForest $R^2$ & PotentialNet $R^2$ \\ \hline 
    Papp & 0.150 (0.081, 0.232) & 0.381 (0.292, 0.468) \\ \hline
    logD & 0.394 (0.305, 0.480) & 0.603 (0.528, 0.670) \\ \hline
    \end{tabular}
\end{table*}

\begin{table*}[ht]
	\caption{Literature\cite{over2016structural} Molecules for which performance improvement of PotentialNet with graph features over Random Forest with pair features is greatest}
    \label{nchembio_mols}
    \begin{tabular}{l l l l l l l}
    \hline
   Molecule & Property & Exp. & RandomForest  & PotentialNet & SMILES \\ \hline
   		 \includegraphics[trim=0 0 0 -5,scale=0.3]{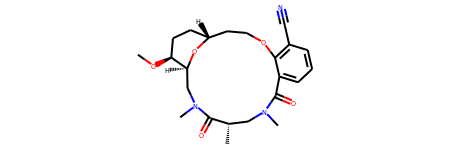} & Papp & 2.33 & 74.75 & 4.90 & \shortstack{\tiny{CO[C@H]1CC[C@H]2CCOc3ccc}\\\tiny{(cc3C(=O)N(C)CCCC(=O)N(C)C[C@@H]1O2)C\#N}} \\ \hline
		 \includegraphics[trim=0 0 0 -5,scale=0.3]{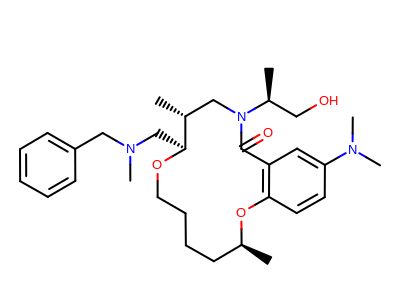} & Papp & 74.34 & 21.51 & 52.42 & \shortstack{\tiny{C[C@@H(CO)N1C[C@@H(C[C@@H(CN(C)Cc2ccccc2)} \\ \tiny{OCCCC[C@@H(C)Oc3ccc(cc3C1=O)N(C)C}} \\ \hline
		 \includegraphics[trim=0 0 0 -5,scale=0.3]{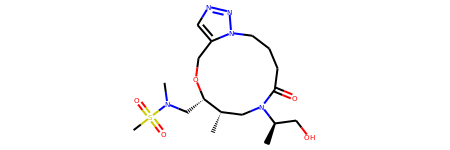} & logD & -0.70 & 1.93 & 0.34 & \shortstack{\tiny{C[C@H](CO)N1C[C@H](C)[C@H]} \\ \tiny{(CN(C)S(=O)(=O)C)OCc2cnnn2CCCC1=O}} \\ \hline 
		 \includegraphics[trim=0 0 0 -5,scale=0.3]{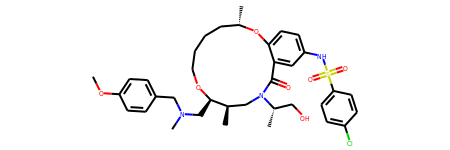} & logD & 4.70 & 2.63 & 3.74 & \shortstack{\tiny{COc1ccc(CN(C)C[C@H]2OCCCC[C@H]C)Oc3ccc(NS(=O)} \\ \tiny{(=O)c4ccc(Cl)cc4)cc3C(=O)N(C[C@H]2C)[C@@H](C)CO)cc1}}
    \end{tabular}
\end{table*}

\section{Figures}

\clearpage

\subsection{Performance of models trained on Merck data on held out literature macrocycle data}

\begin{figure*}[!htb]
\caption{Scatter plots of predictions by models pre-trained on Merck data with both PotentialNet and Random Forest vs. Experiment for logD Measurements on Macrocycles in ref \cite{over2016structural}}
\includegraphics[width=6in]{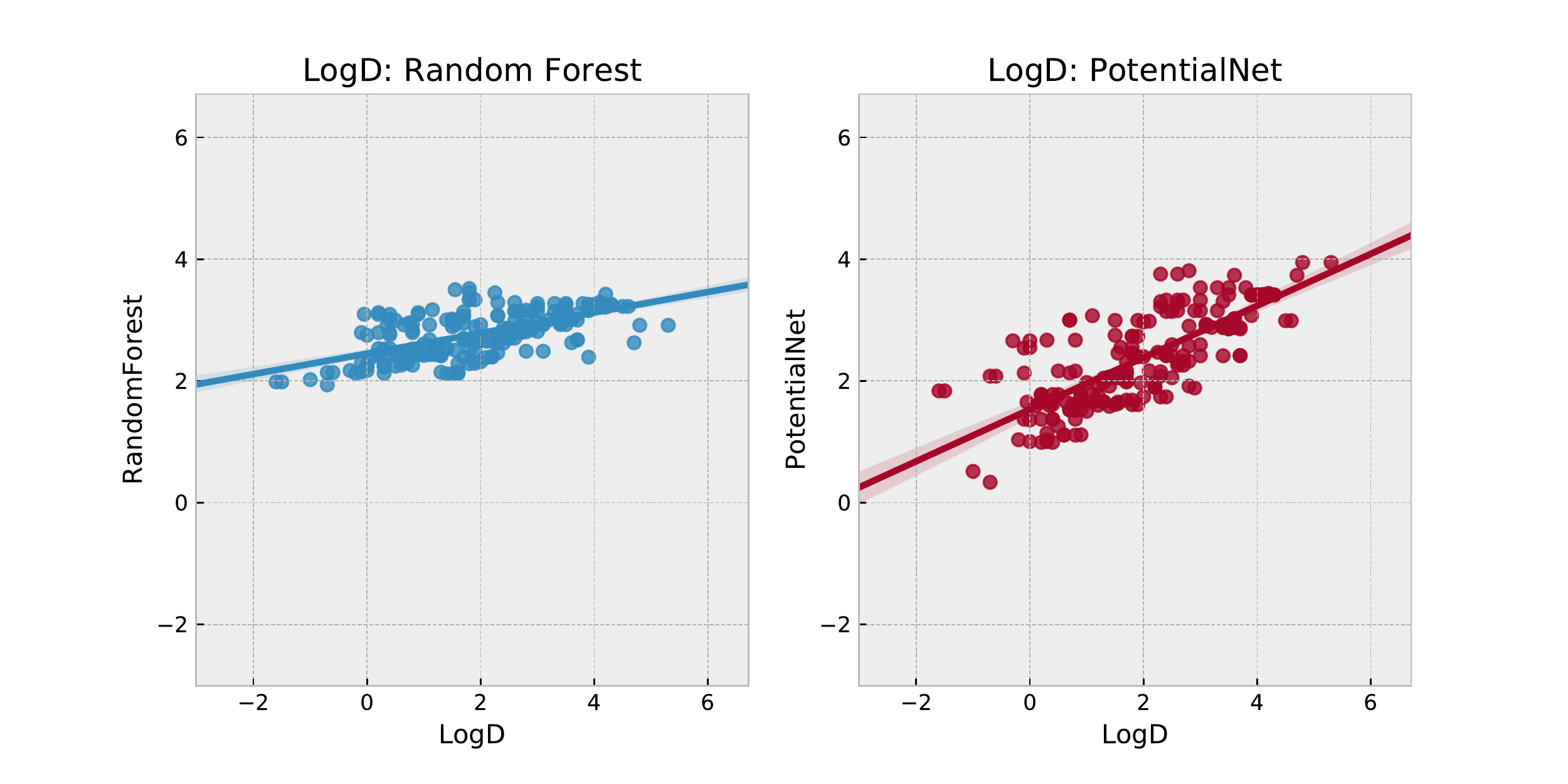}
\end{figure*}

\clearpage

\section*{Acknowledgments}
We thank Juan Alvarez, Andy Liaw, Matthew Tudor, Isha Verma, and Yuting Xu for their helpful comments and insightful discussion in preparing this manuscript. 

\section*{References}

\bibliography{library}

\pagebreak
\widetext
\clearpage

\begin{center}
\textbf{\large Supplemental Figures}
\end{center}
%%%%%%%%%% Merge with supplemental materials %%%%%%%%%%
%%%%%%%%%% Prefix a "S" to all equations, figures, tables and reset the counter %%%%%%%%%%
\setcounter{equation}{0}
\setcounter{figure}{0}
\setcounter{table}{0}
\setcounter{page}{1}
\makeatletter
\renewcommand{\theequation}{S\arabic{equation}}
\renewcommand{\thefigure}{S\arabic{figure}}
\renewcommand{\bibnumfmt}[1]{[S#1]}
\renewcommand{\citenumfont}[1]{S#1}
%%%%%%%%%% Prefix a "S" to all equations, figures, tables and reset the counter %%%%%%%%%%

\pagebreak

\maxdeadcycles=1000
%\extrafloats{1000}

\begin{figure*}[!htb]
\caption{Temporal plus Molecular Weight split: Scatter plot of errors for two models: ABS(PotentialNet - Experiment) vs. ABS(Random Forest - Experiment) for Fu,p human}
\label{err_fu}
\includegraphics[width=6in]{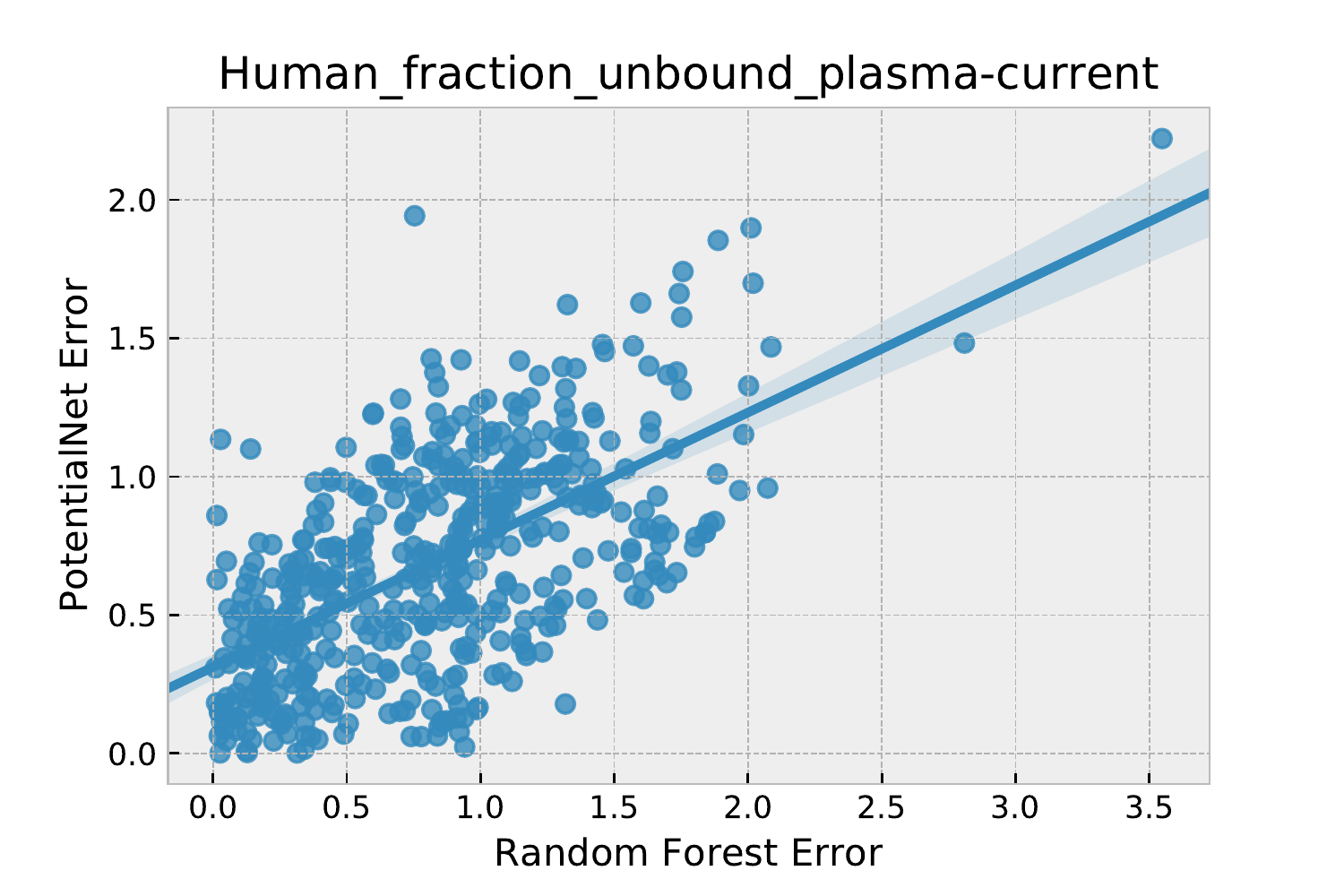}
\end{figure*}

\begin{figure*}[!htb]
\caption{Plot of $R^2$ versus test set size for all datasets for temporal split. Train and test set sizes for each assay for temporal split can be found in SI Figure \ref{temporal_nmols}. Test set size is a qualitatively weak predictor of performance for a deep learning model.}
\includegraphics[width=6in]{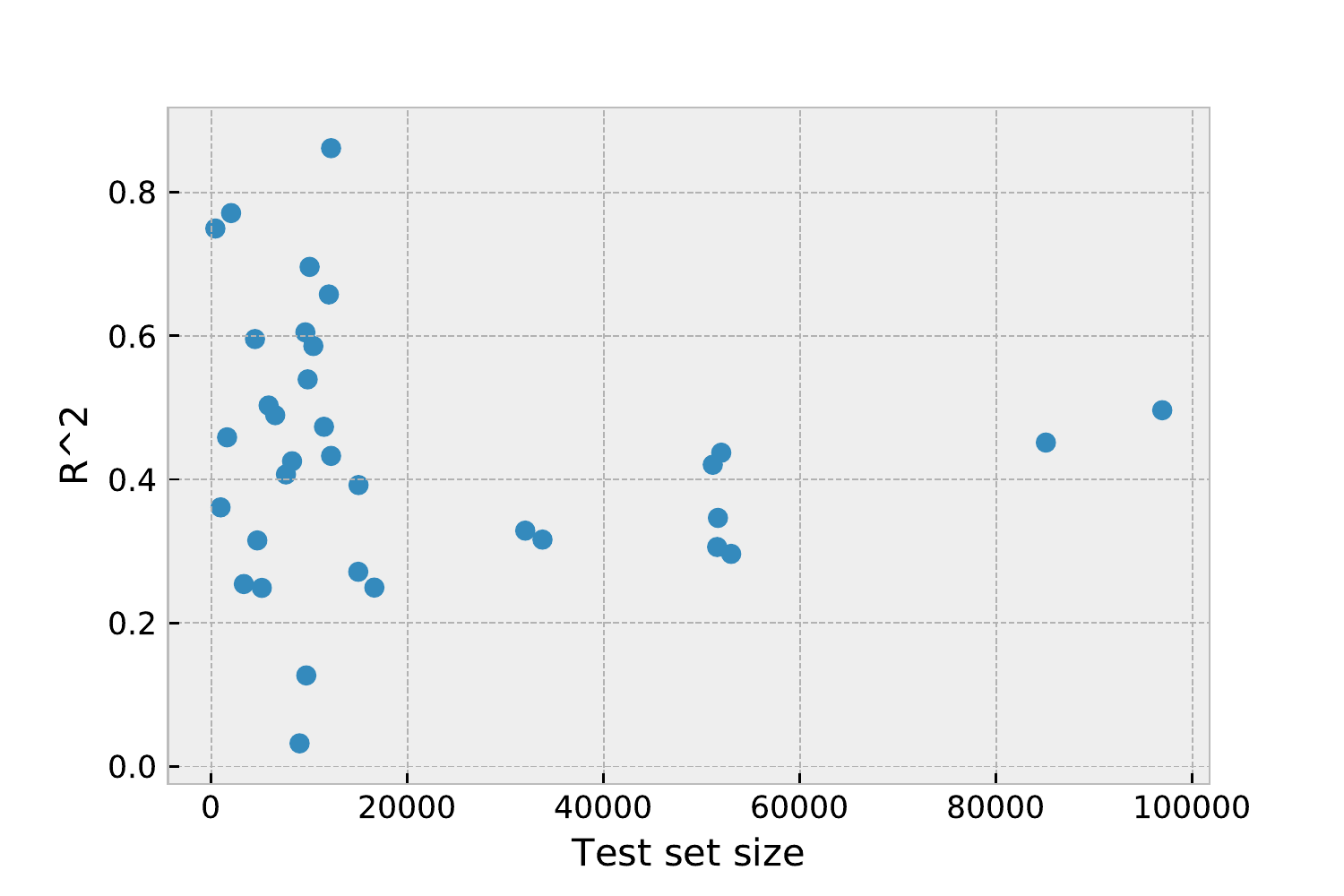}
\end{figure*}

\begin{table*}
\caption{Temporal plus Molecular Weight Split: Aggregate Performance Across all Datasets while only retaining molecules with $MW > 700 \frac{g}{mol}$ in the test set}
\begin{tabular}{lr}
\toprule
{} &         0 \\
\midrule
Mean RandomForest R\textasciicircum2             &    0.091 \\
Mean PotentialNet R\textasciicircum2             &    0.189 \\
Median RandomForest R\textasciicircum2           &    0.063 \\
Median PotentialNet R\textasciicircum2           &    0.178 \\
Mean Absolute R\textasciicircum2 Improvement     &    0.098 \\
Mean Percentage R\textasciicircum2 Improvement   &  318.335 \\
Median Absolute R\textasciicircum2 Improvement   &    0.089 \\
Median Percentage R\textasciicircum2 Improvement &  131.795 \\
\bottomrule
\end{tabular}
\end{table*}

\begin{figure*}[!htb]
\caption{Temporal plus Molecular Weight split for different molecular weight cutoffs. For example, $550$ on the abscissa refers to cross-validation split where \textit{only} molecules with molecular weight $> 550 \frac{g}{mol}$ are retained in the test set. For each cross-validation split listed on the abscissa, only molecules with molecular weight $<500 \frac{g}{mol}$ are retained in the training and validation sets. Therefore, as one moves right along the abscissa, the train-test split becomes more difficult for the machine learner since the gap in molecular weight between train and test molecules becomes larger, forcing the model to do more extrapolation. While performance of both PotentialNet and Random Forests steadily degrades the larger the gap in MW between train and test molecules, PotentialNet also retains an advantage for each Temporal plus Molecular Weight split examined. Median $R^2$ figures are reported over all assays for which there were at least 100 molecules in the test set for a MW cutoff of $700 \frac{g}{mol}$.}
\includegraphics[width=6in]{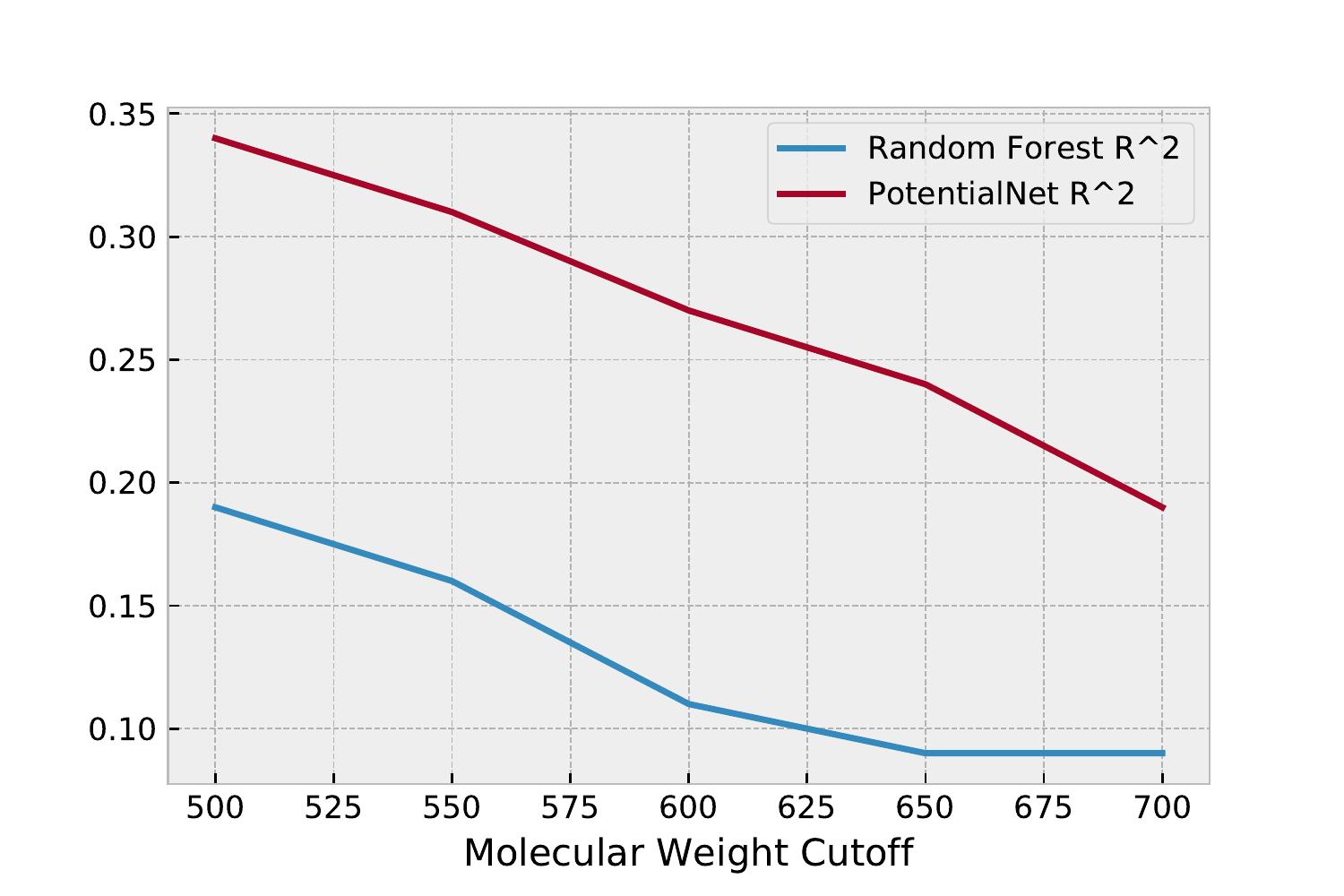}
\end{figure*}

\begin{figure*}[!htb]
\caption{At its core, supervised ML revolves around finding patterns in past data to make predictions on future data. An immediate corollary to this notion is the question of how much past data is needed to make accurate predictions on future data? Here, we performed a training data ablation study. For each of the assay datasets, we temporally removed different proportions of the training and validation data while leaving test data unchanged. For instance, at 60\% data retained on the abscissa, we removed 40\% of the training and validation data points that were temporally or chronologically latest. Therefore, this is equivalent to introducing a time gap between the last training and validation data point and the earliest test data point. Clearly, temporally removing training and validation data systematically diminishes performance of both random forests and of PotentialNet. Nevertheless, the advantage of PotentialNet versus Random Forests in terms of predictive performance remains qualitatively similar regardless of quantity of data removed. It is intriguing that the signal increase from 80\% to 100\% is greater than that of 60\% to 80\%; perhaps this reflects how the most similar ligands to those in the test set are most likely to be temporally adjacent as well.}
\includegraphics[width=6in]{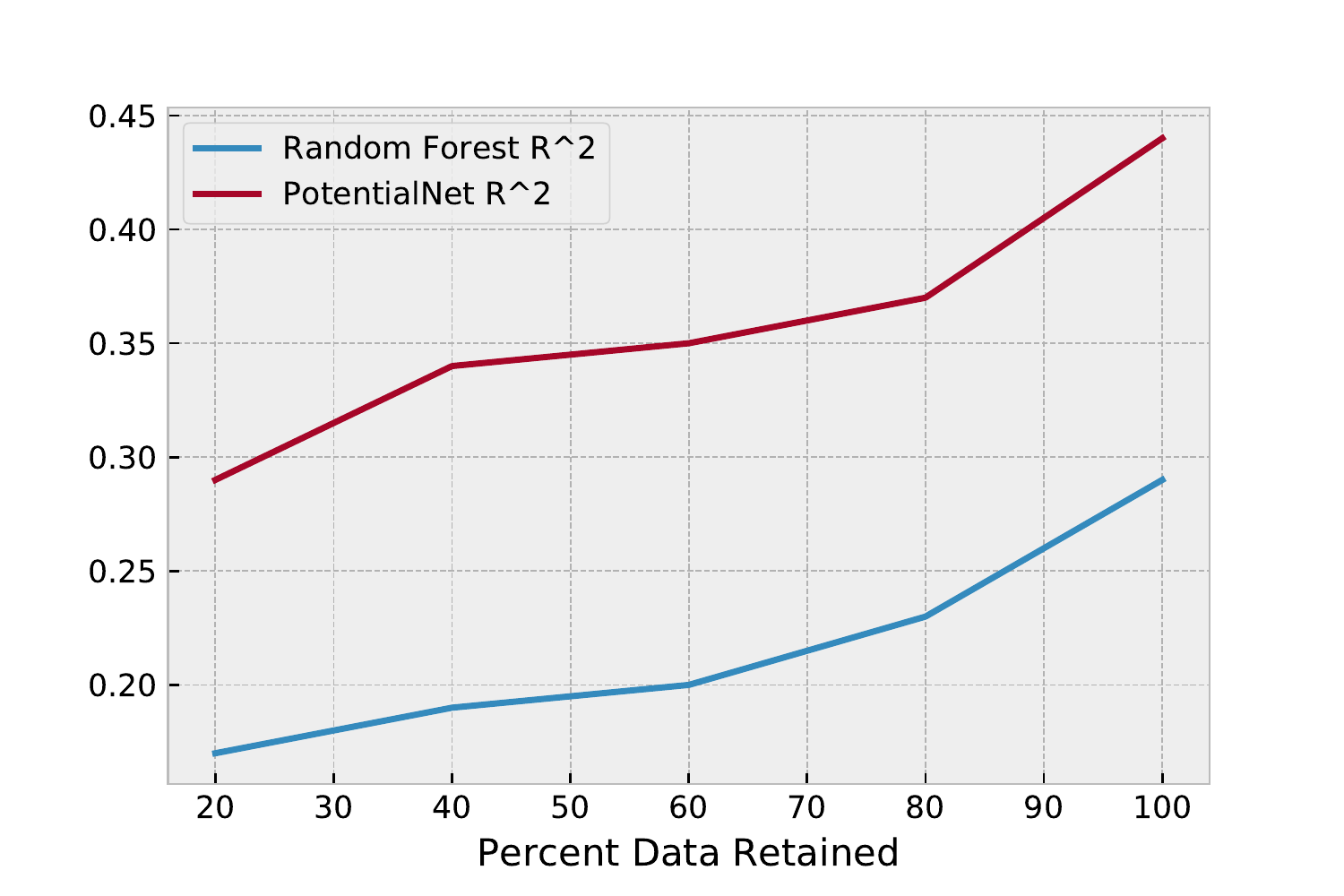}
\end{figure*}

\begin{table*}
\caption{Results on Temporal plus Molecular Weight Split with Random Forests (MIX and Scikit Learn), xgboost, single task multilayer perceptron DNN, single task PotentialNet DNN, and multitask PotentialNet DNN. All results reported with Pearson $R^2$}
\label{st_vs_mt}
\begin{tabular}{lrrrrrr}
\toprule
{} &  RF: MIX &  RF: sklearn &  xgboost &  MLP DNN &  \shortstack{PotentialNet \\ SingleTask} &  \shortstack{PotentialNet \\ MultiTask} \\
\midrule
Rat\_fraction\_unbound\_plasma-current   &    0.012 &        0.097 &    0.054 &    0.483 &                     0.583 &                    0.567 \\
Ca\_Na\_Ion\_Channel\_NaV\_1.5\_Inhibition  &    0.012 &        0.018 &    0.000 &    0.235 &                     0.235 &                    0.247 \\
LOGD                                  &    0.405 &        0.397 &    0.494 &    0.593 &                     0.749 &                    0.651 \\
CLint\_Human\_hepatocyte                &    0.001 &        0.017 &    0.000 &    0.149 &                     0.102 &                    0.029 \\
CYP\_Inhibition\_3A4                    &    0.084 &        0.111 &    0.087 &    0.158 &                     0.228 &                    0.242 \\
PXR\_activation                        &    0.089 &        0.201 &    0.037 &    0.354 &                     0.343 &                    0.364 \\
CYP\_Inhibition\_2D6                    &    0.024 &        0.154 &    0.093 &    0.154 &                     0.136 &                    0.188 \\
\bottomrule
\end{tabular}
\end{table*}

\begin{table*}
\caption{Median $\Delta R^2$ over Random Forests for different ML model types on temporal plus MW split.}
\begin{tabular}{lr}
\toprule
{} &      0 \\
\midrule
(RF: sklearn) - (RF: MIX)              &  0.006 \\
(xgboost) - (RF: MIX)                  & -0.001 \\
(MLP DNN) - (RF: MIX)                  &  0.097 \\
(PotentialNet, SingleTask) - (RF: MIX) &  0.152 \\
(PotentialNet, MultiTask) - (RF: MIX)  &  0.218 \\
\bottomrule
\end{tabular}
\end{table*}

\begin{table*}
\caption{Number of molecules in train and test sets for Temporal plus Molecular Weight split}
\label{temporal_mw_nmols}
\begin{tabular}{lrr}
\toprule
{} &  n\_train\_mols &  n\_test\_mols \\
\midrule
3A4                                   &         24904 &          855 \\
Absorption\_Papp                       &         22043 &          387 \\
CLint\_Dog\_hepatocyte                  &          2419 &          100 \\
CLint\_Dog\_microsome                   &          1381 &           50 \\
CLint\_Human\_hepatocyte                &         15503 &          416 \\
CLint\_Human\_microsome                 &         22839 &          387 \\
CLint\_Rat\_hepatocyte                  &         14744 &          316 \\
\bottomrule
\end{tabular}
\end{table*}

\begin{table*}
\label{prospective_nmols}
\caption{Number of molecules in train and test sets for Prospective Study}
\begin{tabular}{lrr}
\toprule
{} &  n\_train\_mol &  n\_test\_mol \\
\midrule
CYP2C9 IC50          &       206932 &        2483 \\
CYP2D6 IC50          &       206565 &        2333 \\
CYP3A4 IC50          &       204801 &        4121 \\
CaV 1.2 IC50         &       135403 &        3304 \\
Cl, dog              &        18875 &         295 \\
Cl, rat              &        60190 &        1451 \\
Fu,p human           &        17852 &          21 \\
Fu,p rat             &        38583 &        1494 \\
HPLC EPSA            &         8169 &        1844 \\
HPLC logD            &        48914 &       11280 \\
NaV 1.5 IC50         &       128359 &        3356 \\
PAMPA                &         1700 &         300 \\
PGP efflux, human    &        26104 &         169 \\
PGP efflux, rat      &        23402 &          29 \\
PXR activation       &       208309 &        5350 \\
Papp, LLC-PK         &        40131 &         390 \\
Vdss, rat (L/kg)     &        60307 &        1451 \\
hERG inh (MK499)     &       340680 &        4011 \\
hepatocyte Cl, dog   &         6474 &         243 \\
hepatocyte Cl, human &        33251 &        1249 \\
hepatocyte Cl, rat   &        30511 &        1133 \\
microsome Cl, dog    &         3826 &          13 \\
microsome Cl, human  &        41922 &         849 \\
microsome Cl, rat    &        39351 &         824 \\
solubility in FASSIF &       212259 &       10690 \\
t1/2, dog            &        20720 &         332 \\
t1/2, rat            &        66766 &        1490 \\
\bottomrule
\end{tabular}
\end{table*}

\begin{table*}
\caption{Results on Temporal Split with Random Forests (sklearn implementation), single task PotentialNet DNN, and multitask PotentialNet DNN. All results reported with Pearson $R^2$}
\label{st_vs_mt_temporal}
\begin{tabular}{lrrr}
\toprule
{} &  Random Forest R\textasciicircum2 &  PotentialNet, SingleTask &  PotentialNet, MultiTask \\
\midrule
3A4                                   &              0.450 &                     0.630 &                    0.658 \\
Absorption\_Papp                       &              0.520 &                     0.641 &                    0.696 \\
CLint\_Dog\_hepatocyte                  &              0.298 &                     0.378 &                    0.459 \\
CLint\_Dog\_microsome                   &              0.105 &                     0.119 &                    0.361 \\
CLint\_Human\_hepatocyte                &              0.265 &                     0.390 &                    0.431 \\
CLint\_Human\_microsome                 &              0.382 &                     0.553 &                    0.588 \\
CLint\_Rat\_hepatocyte                  &              0.237 &                     0.354 &                    0.408 \\
\bottomrule
\end{tabular}
\end{table*}

\begin{table*}
\caption{Median $\Delta R^2$ over Random Forests for different ML model types on temporal split.}
\label{temporal_nmols}
\begin{tabular}{lr}
\toprule
{} &      0 \\
\midrule
(PotentialNet, SingleTask) - (Random Forest R\textasciicircum2) &  0.108 \\
(PotentialNet, MultiTask) - (Random Forest R\textasciicircum2)  &  0.152 \\
\bottomrule
\end{tabular}
\end{table*}

\begin{table*}
\caption{Number of molecules in train and test sets for temporal split}
\begin{tabular}{lrr}
\toprule
{} &  n\_train\_mols &  n\_test\_mols \\
\midrule
3A4                                   &         36004 &        12001 \\
Absorption\_Papp                       &         30097 &        10032 \\
CLint\_Dog\_hepatocyte                  &          4854 &         1618 \\
CLint\_Dog\_microsome                   &          2868 &          956 \\
CLint\_Human\_hepatocyte                &         24937 &         8312 \\
CLint\_Human\_microsome                 &         31440 &        10480 \\
CLint\_Rat\_hepatocyte                  &         22882 &         7627 \\
\bottomrule
\end{tabular}
\end{table*}

\begin{table*}
\caption{Results on select Merck KAGGLE datasets}
\begin{tabular}{lrlrlrr}
\toprule
{} Dataset &  \shortstack{RandomForest \\ R\textasciicircum2} & \shortstack{RandomForest \\R\textasciicircum2, 95\% CI} &  \shortstack{PotentialNet \\ R\textasciicircum2} & \shortstack{PotentialNet \\R\textasciicircum2, 95\% CI} &  \shortstack{Absolute\\Improvement} & \shortstack{Percentage\\ Improvement} \\
\midrule
2C8                              &             0.178 &           (0.165, 0.191) &             0.296 &            (0.281, 0.31) &                 0.118 &                  66.128 \\
2C9BIG                           &             0.304 &            (0.298, 0.31) &             0.424 &            (0.419, 0.43) &                 0.120 &                  39.421 \\
2D6                              &             0.148 &           (0.139, 0.158) &             0.265 &           (0.253, 0.276) &                 0.116 &                  78.123 \\
3A4small           &             0.434 &           (0.423, 0.445) &             0.629 &            (0.62, 0.637) &                 0.195 &                  44.894 \\
ANRINA                           &             0.014 &           (0.003, 0.033) &             0.047 &           (0.024, 0.077) &                 0.033 &                 228.103 \\
BACE                             &             0.636 &           (0.622, 0.651) &             0.638 &           (0.624, 0.652) &                 0.002 &                   0.304 \\
CAV                              &             0.400 &           (0.389, 0.411) &             0.503 &           (0.493, 0.513) &                 0.103 &                  25.761 \\
\bottomrule
\end{tabular}
\end{table*}

\begin{table*}
\caption{Aggregate Results on select Merck KAGGLE datasets}
\begin{tabular}{lr}
\toprule
{} &        0 \\
\midrule
Mean RandomForest R\textasciicircum2             &   0.343 \\
Mean PotentialNet R\textasciicircum2             &   0.405 \\
Median RandomForest R\textasciicircum2           &   0.318 \\
Median PotentialNet R\textasciicircum2           &   0.416 \\
Mean Absolute R\textasciicircum2 Improvement     &   0.062 \\
Mean Percentage R\textasciicircum2 Improvement   &  34.638 \\
Median Absolute R\textasciicircum2 Improvement   &   0.069 \\
Median Percentage R\textasciicircum2 Improvement &  22.033 \\
\bottomrule
\end{tabular}
\end{table*}

\begin{table*}[!htb]
\caption{Temporal split: Spearman Rho for Random Forest and PotentialNet predictions}
\begin{tabular}{lrlrlrr}
\toprule
{} Dataset &  \shortstack{RandomForest \\ Rho} & \shortstack{RandomForest \\Rho, 95\% CI} &  \shortstack{PotentialNet \\ Rho} & \shortstack{PotentialNet \\Rho, 95\% CI} &  \shortstack{Absolute\\Improvement} & \shortstack{Percentage\\ Improvement} \\
\midrule
t1/2, dog            &             0.445 &           (0.423, 0.467) &             0.447 &           (0.425, 0.468) &                 0.002 &                   0.341 \\
HPLC EPSA            &             0.889 &           (0.879, 0.898) &             0.891 &             (0.882, 0.9) &                 0.002 &                   0.248 \\
Cl, dog              &             0.526 &           (0.505, 0.547) &             0.544 &           (0.524, 0.564) &                 0.018 &                   3.462 \\
Cl, rat              &             0.478 &           (0.466, 0.491) &             0.510 &           (0.498, 0.521) &                 0.031 &                   6.524 \\
Vdss, rat (L/kg)     &             0.583 &           (0.573, 0.594) &             0.620 &             (0.61, 0.63) &                 0.037 &                   6.264 \\
PAMPA                &             0.757 &           (0.713, 0.795) &             0.820 &           (0.786, 0.849) &                 0.063 &                   8.279 \\
MRT, rat (h)         &             0.284 &           (0.266, 0.303) &             0.353 &            (0.335, 0.37) &                 0.068 &                  23.936 \\
\bottomrule
\end{tabular}
\end{table*}

\begin{table*}[!htb]
\caption{Temporal split: spearman's rho, aggregate results}
\begin{tabular}{lr}
\toprule
{} Metric &        Value \\
\midrule
Mean RandomForest Rho             &   0.5096 \\
Mean PotentialNet Rho             &   0.6354 \\
Median RandomForest Rho           &   0.4820 \\
Median PotentialNet Rho           &   0.6590 \\
Mean Absolute Rho Improvement     &   0.1256 \\
Mean Percentage Rho Improvement   &  28.4795 \\
Median Absolute Rho Improvement   &   0.1180 \\
Median Percentage Rho Improvement &  24.7790 \\
\bottomrule
\end{tabular}
\end{table*}

\begin{table*}[!htb]
\caption{Temporal plus MW split: Spearman Rho for Random Forest and PotentialNet predictions}
\begin{tabular}{lrlrlrr}
\toprule
{} Dataset &  \shortstack{RandomForest \\ Rho} & \shortstack{RandomForest \\Rho, 95\% CI} &  \shortstack{PotentialNet \\ Rho} & \shortstack{PotentialNet \\Rho, 95\% CI} &  \shortstack{Absolute\\Improvement} & \shortstack{Percentage\\ Improvement} \\
\midrule
hepatocyte Cl, rat   &             0.421 &           (0.326, 0.507) &             0.065 &          (-0.046, 0.174) &                -0.356 &                 -84.643 \\
Cl, dog              &             0.210 &           (0.085, 0.328) &             0.080 &          (-0.047, 0.204) &                -0.130 &                 -62.032 \\
Cl, rat              &             0.224 &            (0.17, 0.277) &             0.133 &           (0.077, 0.188) &                -0.092 &                 -40.800 \\
hepatocyte Cl, human &             0.204 &            (0.11, 0.294) &             0.154 &           (0.058, 0.246) &                -0.050 &                 -24.634 \\
solubility in FASSIF &             0.431 &           (0.406, 0.456) &             0.388 &           (0.362, 0.414) &                -0.043 &                  -9.935 \\
CYP2D6 IC50          &             0.388 &           (0.357, 0.419) &             0.400 &            (0.368, 0.43) &                 0.011 &                   2.899 \\
PGP efflux, human    &             0.313 &           (0.179, 0.435) &             0.346 &           (0.215, 0.465) &                 0.033 &                  10.638 \\
\bottomrule
\end{tabular}
\end{table*}

\begin{table*}[!htb]
\caption{Temporal plus MW split: spearman's rho, aggregate results}
\begin{tabular}{lr}
\toprule
{} Metric &        Value \\
\midrule
Mean RandomForest Rho             &    0.3149 \\
Mean PotentialNet Rho             &    0.4686 \\
Median RandomForest Rho           &    0.3230 \\
Median PotentialNet Rho           &    0.4570 \\
Mean Absolute Rho Improvement     &    0.1535 \\
Mean Percentage Rho Improvement   &  118.5985 \\
Median Absolute Rho Improvement   &    0.1520 \\
Median Percentage Rho Improvement &   35.2310 \\
\bottomrule
\end{tabular}
\end{table*}

\begin{table*}[!htb]
\caption{Prospective Study: Spearman Rho for Random Forest and PotentialNet Predictions}
\begin{tabular}{lrlrlrr}
\toprule
{} Dataset &  \shortstack{RandomForest \\ Rho} & \shortstack{RandomForest \\Rho, 95\% CI} &  \shortstack{PotentialNet \\ Rho} & \shortstack{PotentialNet \\Rho, 95\% CI} &  \shortstack{Absolute\\Improvement} & \shortstack{Percentage\\ Improvement} \\
\midrule
Cl, dog              &             0.485 &           (0.392, 0.568) &             0.421 &            (0.322, 0.51) &                -0.064 &                 -13.260 \\
t1/2, dog            &             0.600 &           (0.526, 0.664) &             0.559 &            (0.48, 0.629) &                -0.040 &                  -6.746 \\
PGP efflux, rat      &             0.656 &           (0.381, 0.824) &             0.648 &            (0.369, 0.82) &                -0.008 &                  -1.276 \\
Vdss, rat (L/kg)     &             0.653 &           (0.623, 0.682) &             0.654 &           (0.624, 0.683) &                 0.001 &                   0.155 \\
Cl, rat              &             0.573 &           (0.537, 0.606) &             0.580 &           (0.544, 0.613) &                 0.007 &                   1.196 \\
t1/2, rat            &             0.545 &            (0.508, 0.58) &             0.558 &           (0.522, 0.592) &                 0.013 &                   2.452 \\
HPLC EPSA            &             0.832 &           (0.817, 0.846) &             0.862 &            (0.85, 0.874) &                 0.030 &                   3.659 \\
\bottomrule
\end{tabular}
\end{table*}

\begin{table*}[!htb]
\caption{Prospective study: spearman's rho, aggregate results}
\begin{tabular}{lr}
\toprule
{} Metric &        Value \\
\midrule
Mean RandomForest Rho             &   0.5543 \\
Mean PotentialNet Rho             &   0.6449 \\
Median RandomForest Rho           &   0.5520 \\
Median PotentialNet Rho           &   0.6480 \\
Mean Absolute Rho Improvement     &   0.0907 \\
Mean Percentage Rho Improvement   &  18.4518 \\
Median Absolute Rho Improvement   &   0.0920 \\
Median Percentage Rho Improvement &  16.1530 \\
\bottomrule
\end{tabular}
\end{table*}

\clearpage

\begin{figure*}[!htb]
\caption{To use a more traditional metric of molecular similarity to motivate an additional cross-validation procedure, for each assay dataset, we selected newer test molecules with varying maximum Tanimoto similarities to the older training molecules. The training set is identical to that used in temporal split, and the test sets are Tanimoto cutoff-based subsets of the temporal test sets. We only included assays for which there were at least one hundred molecules with a maximum Tanimto similarity less than 0.3 to the training set. Temporal plus Tanimoto split: Aggregate $R^2$ versus Tanimoto similarity cutoff for CYP2C9 Inhibition, CYP2D6 Inhibition, CYP3A4 Inhibition, CaV 1.2 Inhibition, NaV 1.5 Inhibition, PXR Activation, Solubility at pH7, Solubility in FASSIF, and hERG Inhibition.}
\includegraphics[width=6in]{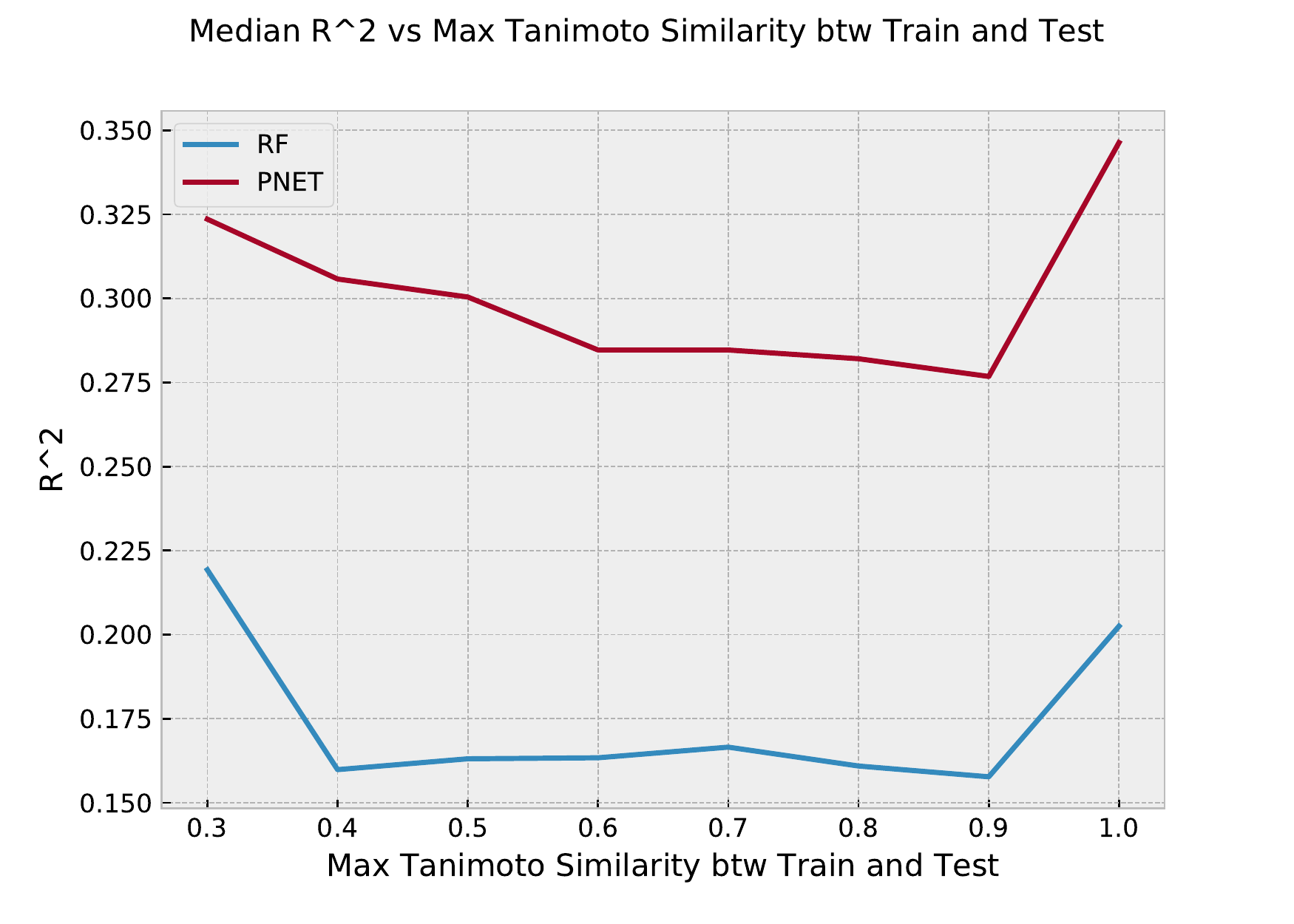}
\end{figure*}

\begin{figure*}[!htb]
\caption{Temporal plus Tanimoto split: Aggregate $R^2$ versus Tanimoto similarity cutoff for hERG Inhibition.}
\includegraphics[width=6in]{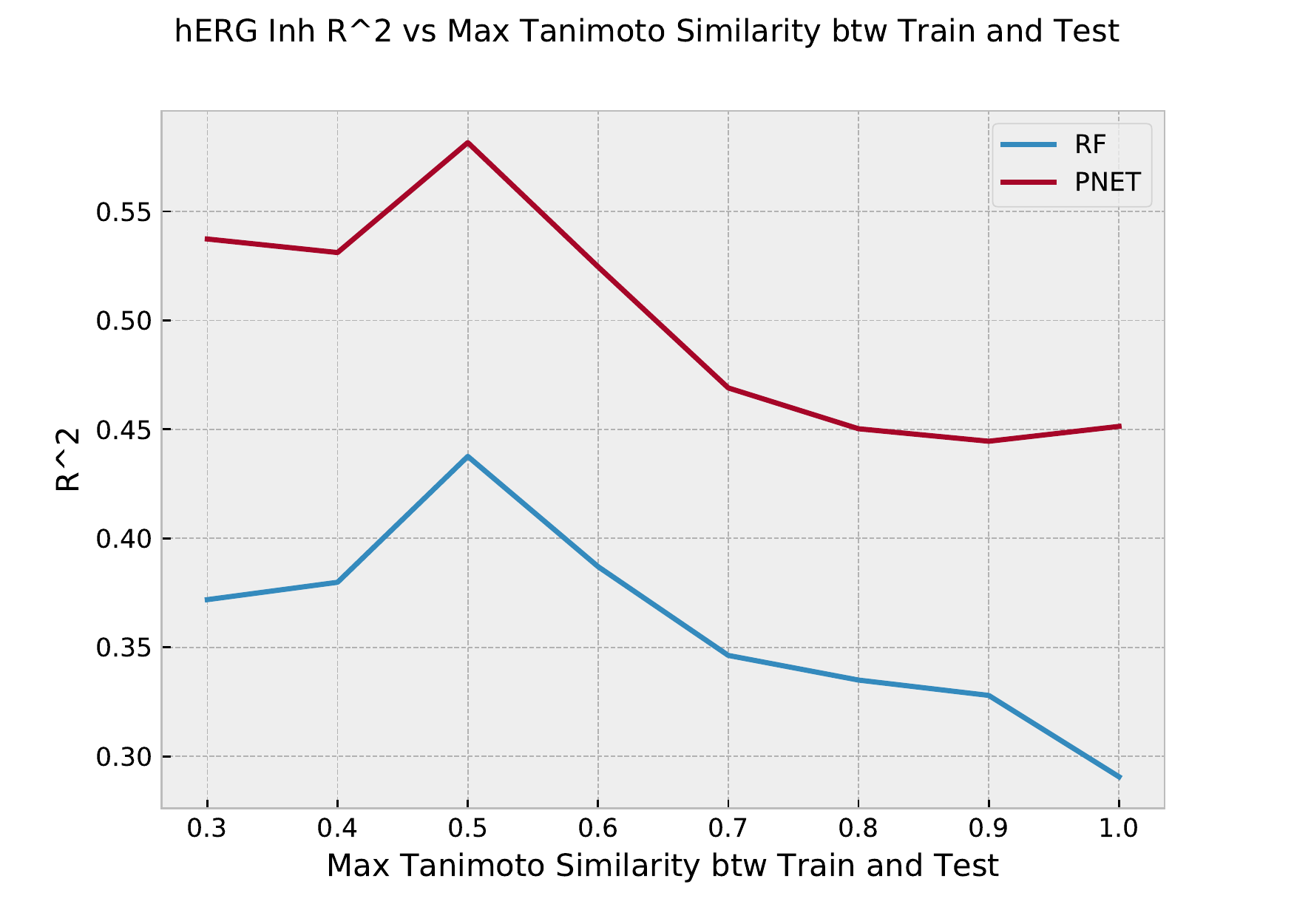}
\end{figure*}

\begin{table*}[!htb]
\caption{Assay names map}
\begin{tabular}{lll}
\toprule
{} &            short name &                                   name w units \\
alt name                             &                       &                                                \\
\midrule
3A4                                  &           CYP3A4 IC50 &                              CYP 3A4 IC50 (uM) \\
Absorption Papp                      &          Papp, LLC-PK &                                                \\
CLint Dog hepatocyte                 &    hepatocyte Cl, dog &                 hepatocyte Cl, dog (ml/min/kg) \\
CLint Dog microsome                  &     microsome Cl, dog &                  microsome Cl, dog (ml/min/kg) \\
CLint Human hepatocyte               &  hepatocyte Cl, human &               hepatocyte Cl, human (ml/min/kg) \\
CLint Human microsome                &   microsome Cl, human &                microsome Cl, human (ml/min/kg) \\
CLint Rat hepatocyte                 &    hepatocyte Cl, rat &                 hepatocyte Cl, rat (ml/min/kg) \\
CLint Rat microsome                  &     microsome Cl, rat &                  microsome Cl, rat (ml/min/kg) \\
CYP Inhibition 2C8                   &           CYP2C8 IC50 &                              CYP 2C8 IC50 (uM) \\
CYP Inhibition 2C9                   &           CYP2C9 IC50 &                              CYP 2C9 IC50 (uM) \\
CYP Inhibition 2D6                   &           CYP2D6 IC50 &                              CYP 2D6 IC50 (uM) \\
CYP Inhibition 3A4                   &           CYP3A4 IC50 &                              CYP 3A4 IC50 (uM) \\
CYP TDI 3A4 Ratio                    &            CYP3A4 TDI &                          CYP 3A4 TDI IC50 (uM) \\
Ca Na Ion Channel\_CaV 1.2 Inhibition &          CaV 1.2 IC50 &                              CaV 1.2 IC50 (uM) \\
Ca Na Ion Channel\_NaV 1.5 Inhibition &          NaV 1.5 IC50 &                              NaV 1.5 IC50 (uM) \\
Clearance Dog                        &               Cl, dog &                            Cl, dog (ml/min/kg) \\
Clearance Rat                        &               Cl, rat &                            Cl, rat (ml/min/kg) \\
EPSA                                 &             HPLC EPSA &                                                \\
Halife Dog                           &             t1/2, dog &                                 t1/2, dog (hr) \\
Halife Rat                           &             t1/2, rat &                                 t1/2, rat (hr) \\
Human fraction unbound plasma        &            Fu,p human &                         PPB, human (\% unbound) \\
LOGD                                 &             HPLC logD &                                           logD \\
MK499                                &      hERG inh (MK499) &                          hERG inh (MK499) (uM) \\
PGP Human 1uM                        &    PGP efflux, humanÿ &                                                \\
PGP Rat 1uM                          &      PGP efflux, ratÿ &                                                \\
PXR activation                       &        PXR activation &  PXR maximum activation relative to rifampicin \\
Rat MRT                              &          MRT, rat (h) &                                                \\
                                     &                       &                                                \\
\bottomrule
\end{tabular}
\end{table*}

\end{document}